\pdfoutput=1

\documentclass[11pt]{article}

\usepackage[preprint]{acl}

\usepackage{times}
\usepackage{latexsym}

\usepackage[T1]{fontenc}

\usepackage[utf8]{inputenc}

\usepackage{microtype}

\usepackage{inconsolata}

\usepackage{graphicx}

\usepackage{booktabs}     
\usepackage{amsmath}      
\usepackage{amssymb}      
\usepackage{multirow}     
\usepackage{ifthen}       
\usepackage{enumitem}      
\usepackage{algorithm}    
\usepackage{algpseudocode}
\usepackage{listings}
\usepackage{subcaption}   
\usepackage{makecell}

\newcommand{\alggen}{algorithm factory}
\newcommand{\thisfolder}{./} 

\usepackage{pifont}
\newcommand{\pifontcheckmark}{\text{\ding{52}}}
\newcommand{\pifontcrossmark}{\text{\ding{56}}}
\newcommand{\good}{\textcolor{teal}{\pifontcheckmark}}
\newcommand{\bad}{\textcolor{magenta}{\pifontcrossmark}}

\usepackage{listings}
\usepackage{tcolorbox}
\usepackage{xcolor}

\definecolor{commentcolor}{RGB}{0,100,0}   
\definecolor{keywordcolor}{RGB}{0,0,255}   
\definecolor{stringcolor}{RGB}{163,21,21}  
\definecolor{identifiercolor}{RGB}{0,0,0}  

\newtcolorbox{mybox}[1][]{
    title=#1,
    fonttitle=\small,
    fontupper=\small,
    left=2mm,
    right=2mm,
    top=1mm,
    bottom=0mm,
}

\lstdefinestyle{mystyle}{
    language=Python,
    commentstyle=\color{commentcolor},
    keywordstyle=\color{keywordcolor}\bfseries,
    stringstyle=\color{stringcolor},
    identifierstyle=\color{identifiercolor},
    basicstyle=\ttfamily\lst@ifdisplaystyle\tiny\fi,
    breakatwhitespace=false,
    breaklines=true,
    captionpos=b,
    keepspaces=true,
    numbers=none,
    numbersep=5pt,
    showspaces=false,
    showstringspaces=false,
    showtabs=false,
    tabsize=2,
    xleftmargin=0pt,
    framexleftmargin=0pt,
}
\usepackage{amsmath}
\usepackage[english]{babel}
\usepackage{hyperref}
\addto\captionsenglish{%
}
\addto\extrasenglish{%
}

\addto\captionsenglish{%
}

\title{Can Large Language Models Invent Algorithms to Improve Themselves?:\\Algorithm Discovery for Recursive Self-Improvement through Reinforcement Learning}

\author{Yoichi Ishibashi \\
  NEC Corporation \\
  \texttt{yoichi-ishibashi@nec.com}
  \\\And
  Taro Yano \\
  NEC Corporation \\
  \texttt{taro\_yano@nec.com} 
  \\\And
  Masafumi Oyamada \\
  NEC Corporation \\
  \texttt{oyamada@nec.com} 
\\}

\begin{document}
\maketitle
\begin{abstract}
Large Language Models (LLMs) have achieved remarkable capabilities, yet their improvement methods remain fundamentally constrained by human design. We present \emph{Self-Developing}, a framework that enables LLMs to autonomously discover, implement, and refine their own improvement algorithms.
Our approach employs an iterative cycle where a seed model generates algorithmic candidates as executable code, evaluates their effectiveness, and uses Direct Preference Optimization to recursively improve increasingly sophisticated improvement strategies. We demonstrate this framework through model merging, a practical technique for combining specialized models.
Self-Developing successfully discovered novel merging algorithms that outperform existing human-designed algorithms. On mathematical reasoning benchmarks, the autonomously discovered algorithms improve the seed model's GSM8k performance by 6\% and exceed human-designed approaches like Task Arithmetic by 4.3\%. Remarkably, these algorithms exhibit strong generalization, achieving 7.4\% gains on out-of-domain models without re-optimization.
Our findings demonstrate that LLMs can transcend their training to invent genuinely novel optimization techniques. This capability represents a crucial step toward a new era where LLMs not only solve problems but autonomously develop the methodologies for their own advancement.
\end{abstract}

\section{Introduction}
The advancement of Large Language Models (LLMs) is having a significant impact on society~\cite{DBLP:conf/nips/VaswaniSPUJGKP17,DBLP:conf/nips/BrownMRSKDNSSAA20,DBLP:journals/corr/abs-2407-21783,guo2025deepseek}.
LLMs have been continuously improved by human experts' knowledge and experience, realizing advanced capabilities such as mathematical reasoning or code generation~\cite{DBLP:journals/corr/abs-2303-08774}.
Building on these advanced capabilities, researchers are increasingly focusing on developing self-improving methods for LLMs to autonomously improve their performance without human intervention, with the goal of automating the LLM development process itself.
Research on self-improvement of LLMs includes approaches such as fine-tuning using self-generated data~\cite{DBLP:conf/icml/YuanPCLSXW24,DBLP:journals/corr/abs-2308-08998,DBLP:journals/corr/abs-2406-03816,DBLP:conf/acl/WangKMLSKH23,DBLP:conf/iclr/XuSZG0FTLJ24}, self-play~\cite{DBLP:journals/corr/abs-2401-05654,DBLP:journals/corr/abs-2404-10642}, and planning using feedback from environment~\cite{DBLP:conf/nips/ShinnCGNY23,DBLP:conf/nips/MadaanTGHGW0DPY23}. 
However, a fundamental limitation is that the exploration of the strategies to improve LLMs  (model-improving algorithms) remains constrained by human knowledge and imagination.

\begin{figure}[t]
    \centering
    \includegraphics[clip, width=6.6cm]{\thisfolder 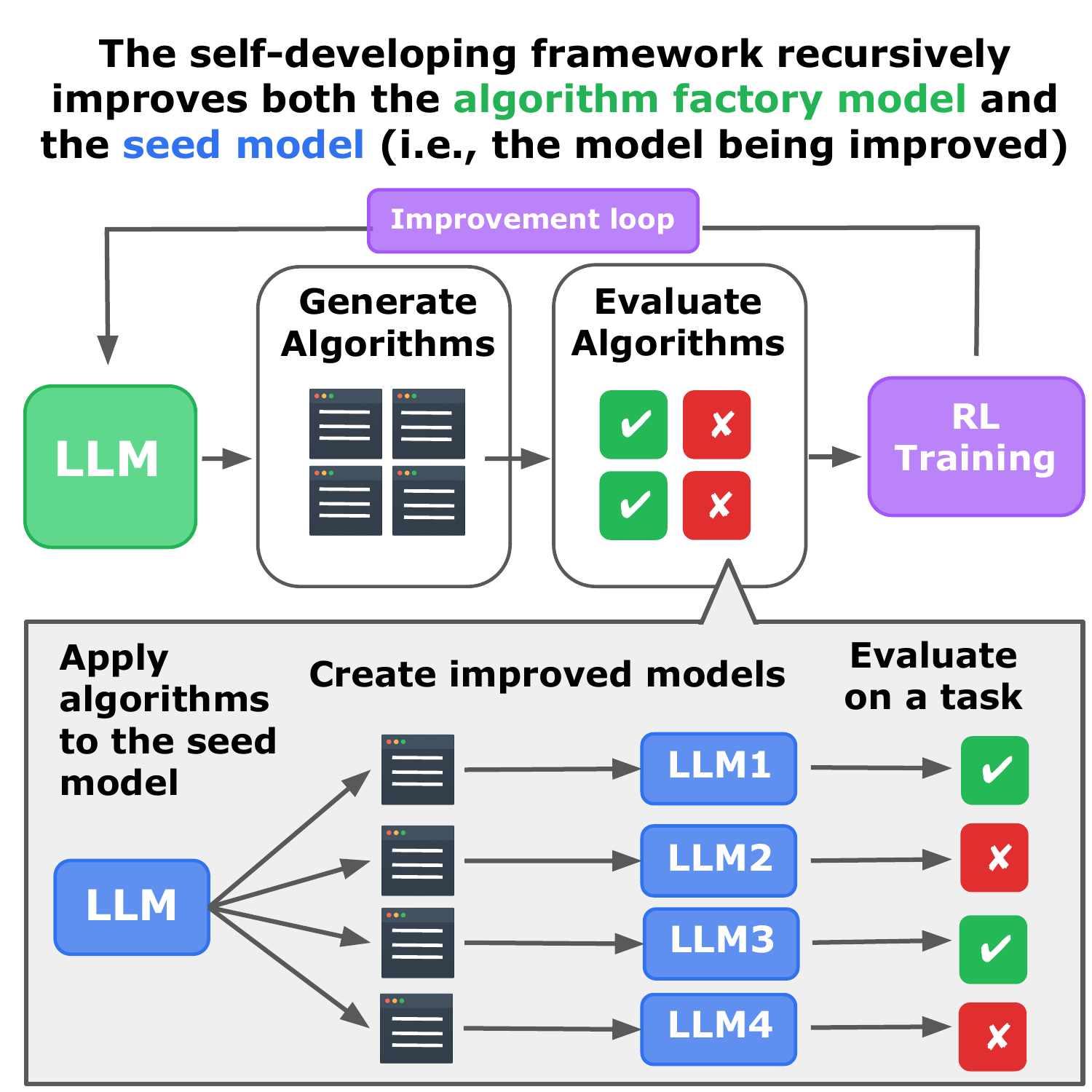}
    \caption{\textbf{The concept of our Self-Developing framework.} The algorithm factory (LLM) generates model-improvement algorithms as executable code and evaluates their effectiveness (\good/\bad). High-performing algorithms serve as positive examples and low-performing ones as negative examples to train the algorithm factory via reinforcement learning, enabling it to generate better algorithms in subsequent iterations. Simultaneously, these improved algorithms are applied to the seed model to create enhanced variants, thus improving both the algorithm generation capability and the target models.}
    \label{fig:concept}
\end{figure}

\begin{figure*}[t]
    \centering
    \includegraphics[clip, width=16cm]{\thisfolder 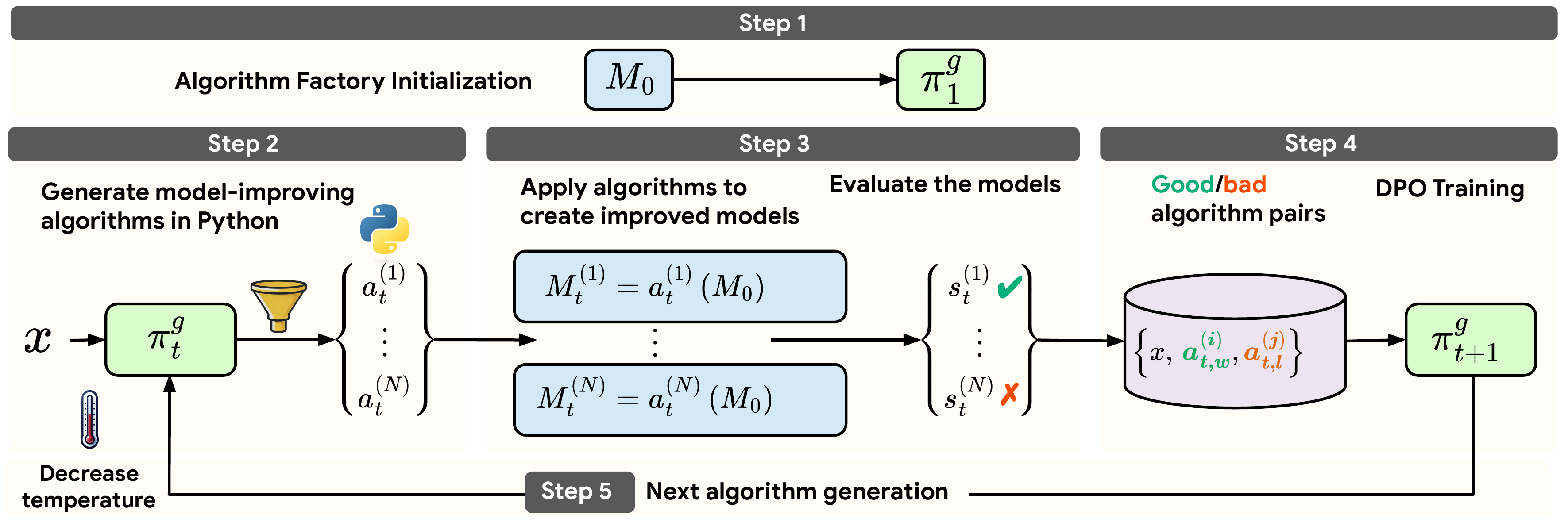}
    \caption{The overview of \textbf{Self-Developing}. This framework involves the simultaneous improvement of the seed model and its self-improvement algorithms by repeating the following steps: First, the algorithm factory $\pi_t^g$ is initialized by seed model $M_0$ (Step 1). In $t$-th iteration, the algorithm factory $\pi_t^g$ takes a prompt $x$ and generates Python code for model-improvement algorithms (Step 2). Then we apply generated algorithms to the seed model $M_0$ to create improved models. The improved models are evaluated on the target task to measure the algorithm's effectiveness using task scores $s_t^{(i)}$ (Step 3). Based on the scores, preference data consisting of high-performance and low-performance algorithm pairs are created, and the next generation of the algorithm factory $\pi_{t+1}^g$ is trained using DPO (Step 4). 
    In the next iteration, $\pi_{t+1}^g$ is used as an algorithm factory (Step 5).
    }
    \label{fig:overview}
\end{figure*}

Regarding this, as an extreme form of self-improvement, one can ask a question: \emph{Could we empower LLMs to autonomously discover and develop algorithms to improve themselves?} This approach could potentially uncover novel, high-performance algorithms beyond human knowledge and imagination, as exemplified by AlphaGo's `\emph{Move 37}'~\cite{DBLP:journals/nature/SilverHMGSDSAPL16}, thus expanding the frontiers of AI capabilities beyond the limitations of human-designed algorithms.

In this paper, we propose \emph{Self-Developing}, an LLM-based framework that invents model-improving algorithms without the use of human expertise or feedback from external stronger models. 
While our framework can be applied to various types of model improvement algorithms, in this study we specifically focus on model merging algorithms~\cite{DBLP:conf/iclr/IlharcoRWSHF23} that create a single improved model from multiple input models, as a concrete instance of model-improving algorithms.
Our approach iteratively refines two components: the seed model, which is improved using LLM-generated algorithms, and the algorithm factory, which generates these algorithms.

In experiments on mathematical reasoning, Self-Developing invents new model-improving algorithms, which can be considered novel model merging strategies. 
On the GSM8k task, LLM-discovered algorithms surpass human-designed methods such as Task Arithmetic~\cite{DBLP:conf/iclr/IlharcoRWSHF23}, enhancing the seed model by 6\% and outperforming human-designed algorithms by 4.3\%.
Furthermore, the discovered algorithms demonstrate strong transferability to out-of-domain models not used in algorithm generation, surpassing the performance of Task Arithmetic optimized for these models by 7.4\%. Notably, our experiments reveal that the iterative refinement of both the seed model and the algorithm factory plays a crucial role in generating increasingly effective algorithms.

\section{Self-Developing: Learning to Generate Model-Improvement Algorithms}\label{sec:method}
The main objective of this research is to enable LLMs to autonomously generate and apply model-improvement algorithms. Specifically, we address the following challenge:
Given a \emph{seed model} $M_0$, our aim fit to generate models that exceeds $M_0$ without guidance from superior teacher models (e.g., GPT-4~\cite{DBLP:journals/corr/abs-2303-08774}) or human intervention. 
This challenging task requires the seed model to devise and implement self-improvement strategies using only its inherent capabilities and knowledge.
Success in this endeavor is defined by the generated model demonstrating higher performance on specified tasks compared to the seed model. To achieve this goal, we propose a framework that iterates through an improvement cycle, as illustrated in \autoref{fig:overview}. 
The cycle consists of the following steps:
\begin{enumerate}
    \item \textbf{Algorithm Factory Initialization: } We initialize an algorithm factory $\pi_1^g$ by cloning the seed model $M_0$ (i.e., $\pi_1^g = M_0$). Both the seed model $M_0$ and the algorithm factory $\pi_1^g$ are iteratively enhanced.
    \item \textbf{Algorithm Generation  (\autoref{sec:step_gen_alg}): } In the $t$-th iteration, \alggen $\pi_t^g$ generates model improving algorithms ($a_t^{(1)}, a_t^{(2)}, \dots, a_t^{(N)}$).
    \item \textbf{Algorithm Evaluation (\autoref{sec:step_eval_alg}): } We apply the generated algorithms to $M_0$ to create new models ($M_t^{(1)}, M_t^{(2)}, \dots, M_t^{(N)} $). By evaluating these on target tasks, we can measure the effectiveness of the algorithms.
    \item \textbf{Algorithm Factory Refinement (\autoref{sec:step_dpo}): } Based on the evaluation results of the models created by applying the generated algorithms, we refine algorithm factory $\pi_t^g$. We create preference data from effective and ineffective algorithms and train using DPO. This enables the algorithm factory to acquire the ability to generate superior algorithms.
    \item \textbf{Iterative Improvement (\autoref{sec:step_cycle}): } By repeating this process, we simultaneously improve the quality of the algorithms and the performance of the generated models.
\end{enumerate}

\begin{figure*}[t]
  \centering
\vspace{-10pt}
\begin{mybox}[Example]
\vspace{-9pt}
\lstset{basicstyle=\scriptsize\ttfamily}
\begin{lstlisting}
def merge_models(model_dict, device):
    '''
    Develop and implement a novel algorithm for merging the model weights. Your goal is to create a unique and effective strategy that combines various existing techniques and introduces new approaches to achieve optimal performance. Consider integrating methods such as adaptive weighting, hybrid strategies, or advanced heuristics to create a more innovative merging technique.

    Args:
        model_dict (dict): A dictionary where keys are model names and values are the model weights.
        device (torch.device): The device (CPU or GPU) on which the computation will be performed.
        
    Returns:
        torch.Tensor: The weights of the merged model.
    '''
    # *New* strategies for merging the model weights:
    #  1. Adaptive weighting
    #  2. Weighted mean of model weights
    
    # Convert model weights to tensors
    weights = [model.detach().to(device) for model in model_dict.values()]
    
    # Step 1: Adaptive weighting
    weight_factors = {
        'GAIR/Abel-7B-002': 0.6,
        'SciPhi/SciPhi-Mistral-7B-32k': 0.3, 
        'teknium/OpenHermes-2.5-Mistral-7B': 0.4 
    }
    
    # Step 2: Weighted mean of model weights
    weighted_weights = [w * factor for w, factor in zip(weights, weight_factors.values())]
    merged_weights = torch.mean(torch.stack(weighted_weights, dim=0), dim=0)
    
    return merged_weights
\end{lstlisting}
\vspace{-6pt}
\end{mybox}
\vspace{-9pt}
    \caption{An example of a model-improving algorithm. This is a model merging function that performs a simple weighted sum of task vectors. The input is a dictionary (with model names as keys and their respective task vectors as values). The algorithm factory produces a Python function that returns the merged task vector. We input up to ``\texttt{\# * New * strategies for merging the model weights :\textbackslash n}'' to ensure that the algorithm factory begins its description from the merging strategy.}
  \label{fig:func_example1}
\end{figure*}

\subsection{Algorithm Generation}
\label{sec:step_gen_alg}
The algorithm factory is a language model that generates \emph{model-improving algorithms} in the form of \emph{programming code}, which enhance the performance of the \emph{seed model}. Formally, the algorithm factory $\pi^g_t$ used at iteration $t \geq 1$ takes a prompt $x$ that encourages algorithm generation as input and outputs an algorithm $a$: $a_t \sim \pi^g_t(a \mid x)$. 

A model-improving algorithm $a$ can be arbitrary (Python) code that receives a model $M$ and returns a model $M'$, which is expected to surpass the performance of the original model $M$. For example, a model-improving algorithm might be code that receives a model and adds some random vectors to its weights. Alternatively, a model-improving algorithm might receive multiple models rather than a single seed model and compute the average of those models to generate a robust model. Previously, a lot of work has human-designed such model-improving algorithms, such as Task Arithmetic~\cite{DBLP:conf/iclr/IlharcoRWSHF23}, TIES merging~\cite{DBLP:conf/nips/YadavTCRB23}, and Model Stock~\cite{DBLP:conf/eccv/JangYH24}. In this paper, the proposed algorithm factory aims to generate such model-improving algorithms.

In our method, we use seed model $M_0$ as the base model for merging the task vectors~\cite{DBLP:conf/iclr/IlharcoRWSHF23} of merge candidate models $\{C_j\}_{j=1}^K$, which are fine-tuned on different datasets.
The task vector $\tau_{C_j}$ is defined as the weight difference between the seed model $M_0$ and the merge candidate model $C_j$: $\tau_{C_j} = C_j - M_0$. 
\autoref{fig:func_example1} illustrates an example of a model-improving algorithm. 
This simple algorithm implements a merge strategy using a weighted sum of task vectors in Python.
Formally, given the set of task vectors $\{\tau_{C_j}\}_{j=1}^K$, the model-improving algorithm $a_t$ outputs a merged task vector $\tau_t$:
\begin{equation}
    \tau_t = a_t(\tau_{C_1}, \dots, \tau_{C_K}).
\end{equation}
We obtain a merged model by adding $\tau_t$ to the seed model:
\begin{equation}
    M_t = M_0 + \tau_t.
\end{equation}

\subsection{Algorithm Evaluation}
\label{sec:step_eval_alg}
The objective of the algorithm factory $\pi^g_t$ is to generate algorithms that enhance the seed model's performance on target tasks. However, in the initial iteration, $\pi^g_1$ is untrained and unable to generate effective algorithms. Therefore, in subsequent iterations, we train the algorithm factory to generate more effective algorithms. We evaluate the merged models obtained from the generated algorithms on the target tasks, and based on these evaluations, we create preference data to train the algorithm factory. 

We assess the effectiveness of the algorithm by evaluating the model generated with the algorithm on the target tasks.
First, from the set of algorithms generated as Python functions, we remove those that are non-executable or result in timeouts, obtaining a set of executable algorithms $\{a_t^{(i)}\}_{i=1}^N$. Second, these algorithms are applied to the task vectors $\{\tau_{C_j}\}_{j=1}^K$ and merged with $M_0$ to generate new models $\{M_t^{(i)}\}_{i=1}^N$. Then, we evaluate the new models on the development set of the downstream tasks, and a task score $s_t^{(i)} \in \mathbb{R}$ is calculated for each model. 
These scores indicate the effectivenesses of the algorithms.
The set of evaluation results $\{s_t^{(i)}\}_{i=1}^N$ obtained for all executable algorithms is used to create the preference data as described in \autoref{sec:step_dpo}.

\subsection{Algorithm Factory Refinement}
\label{sec:step_dpo}
To generate increasingly superior algorithms, we train the algorithm factory using Direct Preference Optimization~\cite[DPO;][]{DBLP:conf/nips/RafailovSMMEF23}.
The key to this learning process lies in the utilization of preference data based on the performance of the generated algorithms. We evaluate the set of generated algorithms $\{a_t^{(i)}\}_{i=1}^N$, selecting high-performance algorithms $a_{t,w}^{(i)}$ (\emph{chosen}) and low-performance algorithms $a_{t,l}^{(j)}$ (\emph{rejected}) based on their evaluation scores $\{s_t^{(i)}\}_{i=1}^N$. The preference information $a_{t,w}^{(i)} \succ a_{t,l}^{(j)}$ is then incorporated into the model's learning process. This allows $\pi^g_t$ to learn the characteristics of effective algorithms, thereby enhancing its ability to generate superior algorithms in subsequent iterations.
Specifically, we select the top-ranked and bottom-ranked algorithms based on a threshold and construct the training data as follows:
\begin{equation}
    \mathcal{D} = \{ (x, a_{t,w}^{(i)}, a_{t,l}^{(j)}) \mid s_{t,w}^{(i)} \geq s_{p_w}\ \text{and}\ s_{t,l}^{(j)} \leq s_{p_l} \},
\end{equation}
where $s_{p_w}$ and $s_{p_l}$ represent the score threshold for the top $p_w\%$ and bottom $p_l\%$, respectively. Then, we train the algorithm factory $\pi_t^{g}$ using $\mathcal{D}$ as the preference dataset.

\subsection{Iterative Improvement}
\label{sec:step_cycle}
Our Self-Developing framework focuses on mutually reinforcing the algorithm factory and seed model through iterative learning and evaluation.
In the $(t+1)$-th iteration, we use $\pi_{t+1}^{g}$ as the algorithm factory, which has been trained using DPO from $\pi_{t}^{g}$.
Note that the generated algorithms are always applied to the seed model $M_0$, as the algorithm factory is trained specifically to improve $M_0$.
By repeatedly performing this improvement cycle, the algorithm factory gradually generates more efficient algorithms, creating a cycle where the seed model $M_0$ is simultaneously enhanced along with the evolution of the algorithms.

\section{Main Results}\label{sec:main}
In this section, we demonstrate that Self-Developing can generate algorithms that improve the model itself, and furthermore, these automatically discovered algorithms overperforms the conventional human-designed ones.

\subsection{Setup}
\paragraph{Tasks}
We evaluate our approach using the mathematics-related tasks GSM8k~\cite{DBLP:journals/corr/abs-2110-14168} and MATH~\cite{DBLP:conf/nips/HendrycksBKABTS21}, which have been employed in previous studies~\cite{DBLP:conf/icml/Yu0Y0L24,DBLP:conf/icml/YuanPCLSXW24,DBLP:journals/corr/abs-2404-02893}. For GSM8k, we allocate 100 examples from the test set as a development set and use the remaining 1220 examples as the test set. For MATH, we select 100 examples from each of its 6 subsets (totaling 600 examples) for the development set and use the remaining 4400 examples as the test set. To prevent any indirect leakage of test set information into the training data $\mathcal{D}$ for $\pi_t^g$, we evaluate $\{M^{(i)}_t\}_{i=1}^N$ exclusively on the development set. 
After completing all iterations, we conduct a single evaluation on the test set, focusing on the top 15 models across all iterations that demonstrated the highest performance on the development set.
We perform evaluations using \texttt{lm-evaluation-harness}\footnote{\url{https://github.com/EleutherAI/lm-evaluation-harness}}~\cite{eval-harness}, employing default prompts and few-shot examples. 
For both GSM8k and MATH, we use 5-shot examples and evaluate using Pass@1~\cite{DBLP:journals/corr/abs-2107-03374} with exact match scoring.
During the evaluation process, we use greedy decoding for generating responses.

\paragraph{Models} 
For the seed model $M_0$, we employ \texttt{openchat-3.5-1210}, a fine-tuned variant of \texttt{Mistral-7B-v0.1}\footnote{\url{https://huggingface.co/mistralai/Mistral-7B-v0.1}}~\cite{DBLP:journals/corr/abs-2310-06825}, which has superior code generation capabilities. 
We also select three \texttt{Mistral-7B}-based fine-tuned models as merging candidates: (1) \texttt{Abel-7B-002}\footnote{\url{https://huggingface.co/GAIR/Abel-7B-002}}, which excels in mathematical tasks~\cite{abel}; (2) \texttt{OpenHermes-2.5-Mistral-7B}\footnote{\url{https://huggingface.co/teknium/OpenHermes-2.5-Mistral-7B}}, trained extensively on code instruction data~\cite{OpenHermes2.5}; and (3) \texttt{SciPhi-Mistral-7B-32k}\footnote{\url{https://huggingface.co/SciPhi/SciPhi-Mistral-7B-32k}}, which specializes in scientific domains. These models, fine-tuned for mathematics and science, are expected to enhance the seed model's capabilities~\footnote{While some of these models may show lower performance on specific benchmarks compared to the seed model, they can still contribute to achieving performance beyond individual models when they possess complementary knowledge or different capabilities.}. 
We use \texttt{mergekit}\footnote{\url{https://github.com/arcee-ai/mergekit}}~\cite{DBLP:journals/corr/abs-2403-13257} for model merging, applying the algorithm to task vectors in each MLP layer of Transformer~\cite{DBLP:conf/nips/VaswaniSPUJGKP17}.

\paragraph{Self-Developing}
Our process involves 3 iterations, each generating 3000 algorithms\footnote{After filtering, the number of executable Python functions typically ranged from 100 to 300 in our experiments.}. 
To effectively balance the exploration-exploitation trade-off in iterative DPO, we decrease the temperature in accordance with the progress of the iteration (see \autoref{sec:apdx_temp_decay}).
We set the initial temperature $T_1$ to 1.2 with a decay rate $\beta$ of 0.2, resulting in $T_3 = 0.85$ for the final iteration. 
The prompt $x$, incorporating a one-shot Python implementation example, remains consistent across iterations (see \autoref{sec:prompt}). 
This prompt remains fixed and is used consistently across all iterations.
For DPO, we create preference data $\mathcal{D}$ by selecting the top 3\% ($p_w$) of high-performing algorithms and the bottom 10\% ($p_l$) of low-performing algorithms. 
We reserve 10\% of the training data for development and fine-tune all $\pi_t^g$ linear layers using LoRA~\cite{DBLP:conf/iclr/HuSWALWWC22} with rank $r=256$. We use a learning rate of $1e-6$, $\beta$ of $0.01$, and cap the training at 5000 steps. All experiments run on NVIDIA A100 GPUs. For iterations where $t\geq2$, we augment $\mathcal{D}$ with the top 3 performing algorithms from each preceding iteration $\{a_{1}^{(i)}\}_{i=1}^{N_1}, \cdots, \{a_{t-1}^{(i)}\}_{i=1}^{N_{t-1}}$ (see \autoref{sec:pseudocode}).

\paragraph{Baselines}
We compare our Self-Developing with well-established human-designed model-improving algorithms, selecting representative methods that have demonstrated effectiveness in recent literature. 
Specifically, we include Task Arithmetic~\cite{DBLP:conf/iclr/IlharcoRWSHF23}, TIES Merging~\cite{DBLP:conf/nips/YadavTCRB23}, and Model Stock~\cite{DBLP:conf/eccv/JangYH24} as baselines.
For Task Arithmetic and TIES Merging, we exhaustively evaluate all combinations of mixing ratios of 20\%, 40\%, and 60\% for candidate models for merging on the development set. 
For each task, we select the combination that performs best on the development set and then evaluate this optimal combination on the test set.

\begin{table}[t]
\setlength{\tabcolsep}{0.7mm}
\centering
\scalebox{0.68}[0.68]{
\begin{tabular}{lccc}
\toprule
\textbf{Models}               & \textbf{GSM8k (\%)} & \textbf{MATH (\%)} \\
\midrule
\midrule
\multicolumn{3}{c}{Base Model (Seed Model)} \\
\midrule
\texttt{openchat-3.5-1210} ($M_0$)       & 70.1       & 0.5      \\
\midrule
\multicolumn{3}{c}{Models for Merging} \\
\midrule
\texttt{Abel-7B-002}   ($C_1$) & 64.8       & 3.7      \\
\texttt{OpenHermes-2.5-Mistral-7B}   ($C_2$) & 60.1       & 1.7      \\
\texttt{SciPhi-Mistral-7B-32k}  ($C_3$) & 56.5       & 1.0      \\
\midrule
\multicolumn{3}{c}{Human-Designed Algorithms (Best Performances)} \\
\midrule
Task Arithmetic~\cite{DBLP:conf/iclr/IlharcoRWSHF23}    & 71.9       & \textbf{8.5}      \\
TIES Merge~\cite{DBLP:conf/icml/Yu0Y0L24}  & 71.8      & 8.4      \\
Model Stock~\cite{DBLP:conf/eccv/JangYH24}    & 39.5     & 6.1     \\
\midrule
\multicolumn{3}{c}{LLM-Designed Algorithms (Top 3 Performances)} \\
\midrule
1st (GSM8k:  \autoref{fig:func_example_A}, MATH: \autoref{fig:func_example_I})       & \textbf{76.1}       & \textbf{8.5}   \\
2nd (GSM8k:  \autoref{fig:func_example_B}, MATH: \autoref{fig:func_example_J})       & \textbf{76.1}       & 8.4   \\
3rd (GSM8k:  \autoref{fig:func_example_C}, MATH: \autoref{fig:func_example_K})       & \textbf{76.0}       & 8.4  \\
\bottomrule
\end{tabular}
}
\caption{Performance evaluation results of each method on the GSM8k and MATH tasks. 
The algorithms discovered by Self-Developing outperforms the seed model and existing model-improving algorithms. These results demonstrate that LLMs can invent effective model-improving algorithms that surpass human-designed techniques.}
\label{tab:main_results}
\end{table}

\subsection{Results}
\label{sec:main_results1}
Table \ref{tab:main_results} presents the performance comparison between human-designed algorithms and algorithms discovered by Self-Developing on the GSM8k and MATH tasks.
For our approach, we display the top three performances obtained across all iterations of our algorithm discovery process.

\paragraph{Q1: Can LLMs evolve using self-discovered algorithms?} 
The results in \autoref{tab:main_results} demonstrate that LLMs can improve their own performance using self-discovered model-improvement algorithms. The models applying the top three algorithms discovered by the LLM consistently outperform both the seed model (\texttt{openchat-3.5-1210}) and the three models for merging. Notably, on the GSM8k task, we achieve the highest  accuracy of 76.1\%, representing a significant performance gain of about 6\% over the seed model's 70.1\%. For the MATH task, our best model reaches 8.5\% accuracy, showing a substantial improvement from the seed model's 0.5\%. These results are particularly remarkable considering that powerful external models like GPT-4 were not used in the algorithm generation process.

\paragraph{Q2: Do discovered algorithms surpass human-designed ones?} 
A significant finding is that our proposed method autonomously discovered algorithms that outperform human-designed techniques such as Task Arithmetic and TIES merging. 
As shown in \autoref{tab:main_results}, models created using the LLM-discovered algorithms consistently demonstrate higher performance on the GSM8k task compared to Task Arithmetic (76.1\% vs 71.9\%) and TIES merging (76.1\% vs 71.8\%). 
On the MATH task, our best model is comparable to the top performance of Task Arithmetic (8.5\%) and slightly outperforms TIES merging (8.4\%).
Outperforming Task Arithmetic, renowned for its strength in mathematical reasoning~\cite{DBLP:conf/icml/Yu0Y0L24}, highlights our autonomous algorithm discovery's effectiveness and its potential to surpass well-crafted human-designed algorithms.

\paragraph{Q3: Does training the algorithm factory improve algorithm quality?} 
One of the key contributions of our work is the automatic improvement of model-improving algorithms, which is made possible by training the algorithm factory. Our findings demonstrate that this training leads to a significant enhancement in the quality of generated algorithms, enabling a novel form of LLM self-improvement.
\autoref{tab:iteration_progress} shows a clear improvement in performance across iterations, particularly for the MATH task.
We see substantial performance gain in MATH from 7.0\% to 8.5\%. This iterative improvement confirms our method's ability to continuously self-improve through the discovery of increasingly effective algorithms.

\begin{table}[h]
\centering
\small
\begin{tabular}{ccc}
\toprule
\textbf{Model} & \textbf{GSM8k (\%)} & \textbf{MATH (\%)} \\
\midrule
$M_0$ & 70.1 & 0.5 \\
\midrule
$M_1^{\text{best}}$ & 75.8 & 7.0 \\
$M_2^{\text{best}}$ & 76.0 & 7.5 \\
$M_3^{\text{best}}$ & \textbf{76.1} & \textbf{8.5} \\
\bottomrule
\end{tabular}
\caption{Performance progression on the test data for the top-performing models for each iteration selected on the development data, demonstraining the effectiveness of training algorithm factory iteratively.}
\label{tab:iteration_progress}
\end{table}

\begin{figure}[t] 
    \centering
    \includegraphics[width=\linewidth]{\thisfolder 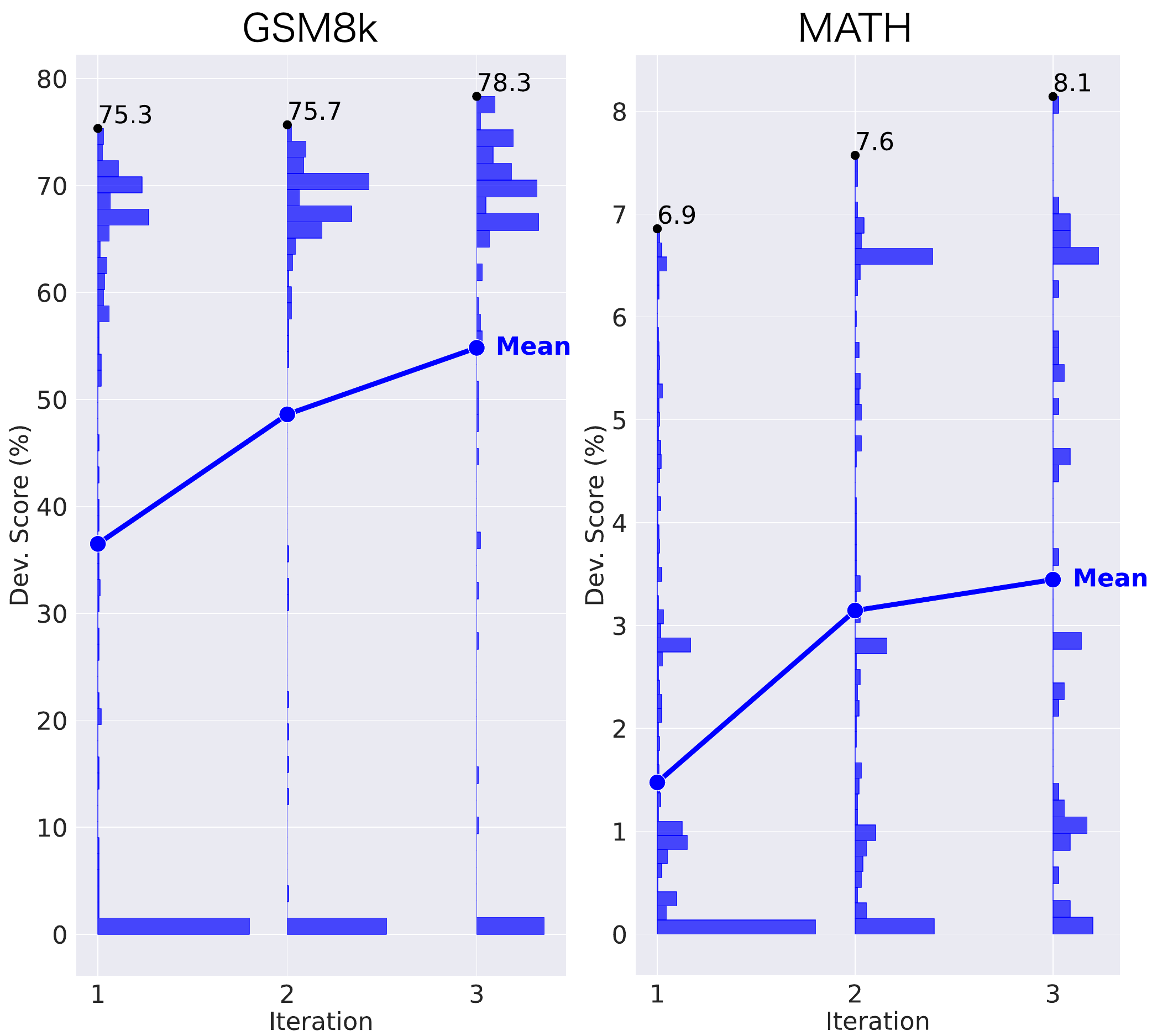}
    \caption{Distribution of algorithm performance on GSM8k and MATH development sets across iterations. Early iterations are dominated by low-performing algorithms, but as learning progresses, the ratio of high-performing algorithms increases significantly.} 
    \label{fig:combined_mean_scores_with_histogram}
\end{figure}

\autoref{fig:combined_mean_scores_with_histogram} demonstrates that the quality of algorithms improves with each iteration. This figure shows the distribution of development scores for models created using algorithms generated in each iteration. In the early iterations, low-performing algorithms were dominant, but as learning progressed, we can observe a significant increase in the ratio of high-performing algorithms.
By training the algorithm factory, the LLM not only discovers effective model-improving algorithms but also refines these algorithms over time, resulting in increasingly enhanced models.

\paragraph{Q4: How Do the Algorithms Evolve Across Iterations?}
Our analysis reveals several interesting characteristics that distinguish LLM-generated algorithms from human-designed ones. These algorithms demonstrate the framework's ability to flexibly combine different techniques, which is particularly evident in their evolution across iterations.

\begin{figure*}[t]
  \centering
\vspace{-10pt}
\begin{mybox}[Discovered Merging Function]
\vspace{-9pt}
\lstset{basicstyle=\scriptsize\ttfamily}
\begin{lstlisting}
def merge_models(model_dict, device):
    # *New* strategies for merging the model weights:
    # 1. Use a combination of weighted averages, element-wise maximums, and element-wise minimums.
    # 2. Apply a Gaussian blur to the weights before merging.
    
    # Convert model weights to tensors and move them to the specified device (CPU or GPU)
    weights = [model.detach().to(device) for model in model_dict.values()]

    def merge_strategy(x, y):
        # Use a combination of weighted averages, element-wise maximums, and element-wise minimums
        return 0.33 * x + 0.33 * y + 0.33 * torch.max(x, y) + 0.01 * torch.min(x, y)

    # Apply a Gaussian blur to the weights before merging
    def gaussian_blur(x, sigma=1.0):
        # Implement the Gaussian blur using convolution
        kernel_size = 2 * sigma + 1
        kernel = torch.Tensor([1 / (2 * sigma * (2 * np.pi) ** 0.5) * torch.exp(-torch.pow(x / (2 * sigma), 2) / (2 * sigma ** 2)) for x in range(-kernel_size // 2, kernel_size // 2)])
        return torch.nn.functional.conv2d(x.unsqueeze(1), kernel.unsqueeze(0), padding=kernel_size // 2)

    # Iteratively merge each subsequent model's weights
    for i, weight in enumerate(weights[1:], start=1):
        weights[0] = merge_strategy(weights[0], weight)
        if i % 5 == 0:
            weights[0] = gaussian_blur(weights[0])

    return weights[0]
\end{lstlisting}
\vspace{-6pt}
\end{mybox}
\vspace{-9pt}
    \caption{A merging algorithm discovered in iteration 1, which combines weighted averages with element-wise maximum and minimum operations. This initial algorithm achieved 71.3\% accuracy in GSM8k development set.}
  \label{fig:func_example_unique_it1}
\end{figure*}

\begin{figure*}[t]
  \centering
\vspace{-10pt}
\begin{mybox}[Discovered Merging Function]
\vspace{-9pt}
\lstset{basicstyle=\scriptsize\ttfamily}
\begin{lstlisting}
def merge_models(model_dict, device):
    # *New* strategies for merging the model weights:
    # 1. Use a combination of weighted averages, element-wise maximums, and element-wise minimums.
    #    - Assign a different weight to each strategy (e.g., 1/3 for averaging, 1/3 for maximum, 1/3 for minimum).
    # 2. Normalize the weight tensors and use a custom distance metric that takes into account both magnitude and direction.
    
    # Convert model weights to tensors and move them to the specified device (CPU or GPU)
    weights = [model.detach().to(device) for model in model_dict.values()]

    def custom_distance(x, y):
        # Calculate L2 norms of x and y
        x_norm = torch.norm(x, 2) + 1e-12
        y_norm = torch.norm(y, 2) + 1e-12
        # Normalize x and y
        x = x / torch.norm(x, 2) + 1e-12
        y = y / torch.norm(y, 2) + 1e-12
        # Compute the custom distance as a weighted sum of L2 distance and cosine distance
        return (torch.norm(x - y, 2) * 0.4 + (1 - torch.norm(x - y, 2) / (torch.norm(x, 2) + torch.norm(y, 2))) * 0.6 * torch.tensor([1.0]))

    # Initialize merged_weights with the first model's weights
    merged_weights = weights[0].clone()
    n = len(weights)

    # Iteratively merge each subsequent model's weights with adaptive weights for each strategy
    alpha_avg, alpha_max, alpha_min = [1. / n] * 3
    for i, weight in enumerate(weights[1:], 1):
        with torch.no_grad():
            dist = custom_distance(merged_weights, weight)
            # Update the adaptive weights based on the distance
            alpha_avg *= (1 / (1 + dist.pow(1. / 3).item()))
            alpha_max *= (1 / (1 + dist.clamp(min=1.).pow(1. / 3).item()))
            alpha_min *= (1 / (1 + (1 - dist.clamp(max=0.).pow(1. / 3)).item()))
        
        # Merge the weights using the adapted alpha values
        merged_weights = alpha_avg * merged_weights + alpha_max * torch.max(merged_weights, weight) + alpha_min * torch.min(merged_weights, weight)

    return merged_weights
\end{lstlisting}
\vspace{-6pt}
\end{mybox}
\vspace{-9pt}
\caption{A merging algorithm discovered in iteration 3, incorporating adaptive weighting mechanisms based on a custom distance metric. Extends the mixture strategy from iteration 1, achieving 73.6\% on GSM8k development set.}
  \label{fig:func_example_unique_it3}
\end{figure*}

In iteration 1, a weighted mixture strategy (\autoref{fig:func_example_unique_it1}) was discovered that combines weighted averages with element-wise maximum and minimum operations, achieving 71.3\% accuracy in GSM8k development set. This initial algorithm demonstrated the framework's ability to explore sophisticated weight combination methods beyond simple averaging. Interestingly, the algorithm factory also proposed incorporating Gaussian blur, a technique commonly used in computer vision, although this function was not activated in our experimental setting with three models.

In iteration 3, the algorithm evolved to incorporate adaptive weighting mechanisms into the previous mixture strategy of weighted averages and element-wise operations (\autoref{fig:func_example_unique_it3}), reaching 73.6\% accuracy in GSM8k development set. The weights are dynamically adjusted based on a custom distance metric that considers both the magnitude and direction of the weight vectors.

These results validate the necessity of our framework: while Task Arithmetic and TIES merging are constrained to predefined weight combinations, our framework's ability to explore diverse algorithmic strategies enables it to discover more effective solutions beyond simple weight optimization.
The discovered algorithms often involve sophisticated operations that would be difficult to achieve through simpler approaches, such as methods based on custom distance metrics that consider both magnitude and direction of weight vectors, and adaptively adjusted weighting strategies based on model similarity.
For detailed analysis, see \autoref{sec:alg_analysis}.

\section{Transferability of Algorithms}
\label{sec:trans_result}
We analyze the effectiveness of the algorithms discovered by the algorithm factory on out-of-domain models that were not used in the algorithm generation process.

\paragraph{Experimental Setup}
To investigate the transferability of LLM-discovered algorithms, we maintained the original seed model $M_0$ while introducing a new set of candidate models ($C_1^{\text{new}}, C_2^{\text{new}}, C_3^{\text{new}}$) and applied the discovered algorithms to these new models.
From models with capabilities similar to the original merge candidates, we selected \texttt{WizardMath-7B-V1.1}\footnote{\url{https://huggingface.co/WizardLMTeam/WizardMath-7B-V1.1}}~\cite{DBLP:conf/iclr/XuSZG0FTLJ24}, \texttt{BioMistral-7B}\footnote{\url{https://huggingface.co/BioMistral/BioMistral-7B}}~\cite{DBLP:conf/acl/LabrakBMGRD24}, and \texttt{Starling-LM-7B-alpha}\footnote{\url{https://huggingface.co/berkeley-nest/Starling-LM-7B-alpha}}~\cite{zhu2024starlingb} as new candidate models for merging. 
Although these models differ from the candidate models used in algorithm generation, they are fine-tuned based on Mistral-7B~\cite{DBLP:journals/corr/abs-2310-06825} and can therefore be merged with the seed model. 
We apply the top 15 algorithms discovered by the algorithm factory (based on their performance on the development set with the original candidate models) to these new models. For comparison, we chose Task Arithmetic, which showed the second-best performance after our proposed method in \autoref{sec:main}, and apply its top 15 mixing ratios (based on development set with the original candidate models) to these new models.

\begin{figure}[t] 
\centering
\includegraphics[width=\linewidth]{\thisfolder 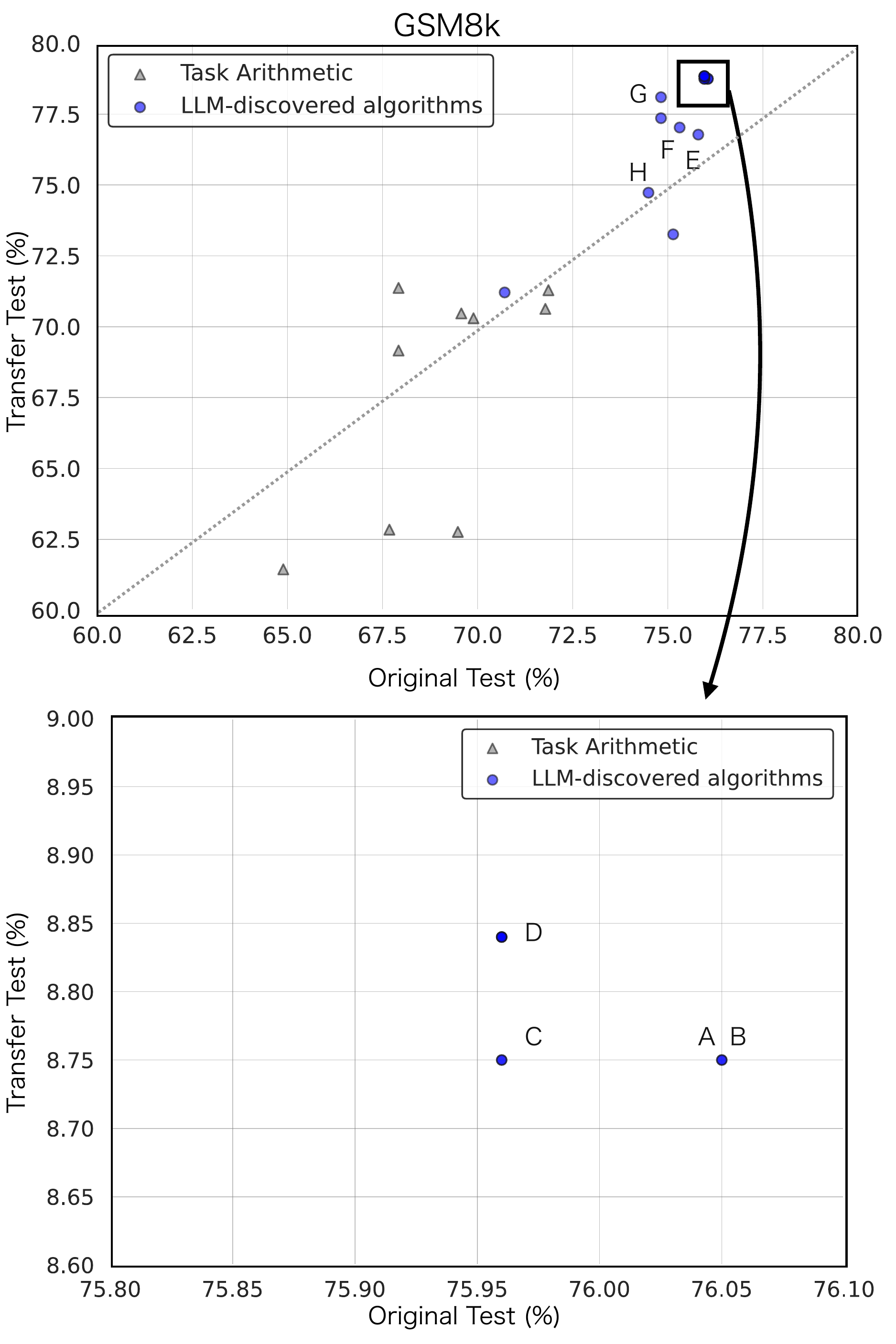} 
\caption{Transferability of the top 15 merge algorithms for the GSM8k task. The x-axis shows the score on the original set of fine-tuned models used for merging, while the y-axis shows the score when the same merging algorithm is applied to a new set of fine-tuned models. Alphabetic labels (A, B, C, etc.) represent discovered algorithms with high transferability, detailed in \autoref{sec:apdx_algorithms}. Points above the diagonal line indicate better transferability, with higher positions showing greater improvement on new models to be merged.}
\label{fig:transferability_gsm8k} 
\end{figure}

\paragraph{Results}
\autoref{fig:transferability_gsm8k} is the results of the transferability evaluation for the algorithms. The algorithms discovered by Self-Developing demonstrated transferability on both GSM8k and MATH tasks. In the GSM8k task, many algorithms maintained high performance even when applied to new candidate models for merging. 
Our LLM-discovered algorithms are positioned above the diagonal line, indicating high scores even when applied to new candidate models. In contrast, the results for Task Arithmetic are concentrated below the diagonal line, suggesting limited transferability.
These findings indicate that the algorithm factory not only generates algorithms optimized for specific model sets but also discovers merge algorithms that maintain high performance on similar candidate models. Similar results are obtained for the MATH task, which are provided in \autoref{sec:apdx_trans}.

\begin{table}[t]
\centering
\scalebox{0.8}[0.8]{
\begin{tabular}{@{}ccc@{}}
\toprule
\textbf{Models}                & \textbf{GSM8k (\%)} & \textbf{MATH (\%)} \\ 
\midrule
\midrule
\multicolumn{3}{c}{Base Model (Seed Model)} \\
\midrule
\texttt{openchat-3.5-1210} ($M_0$)       & 70.1       & 0.5      \\
\midrule
\multicolumn{3}{c}{New Models for Merging} \\
\midrule
\texttt{WizardMath-7B-V1.1}  ($C_1^{\text{new}}$) & 57.4     &  0.03    \\
\texttt{Starling-LM-7B-alpha}  ($C_2^{\text{new}}$) & 75.5      & 0.1     \\
\texttt{BioMistral-7B}  ($C_3^{\text{new}}$) &  0.0      & 0.5     \\
\midrule
\multicolumn{3}{c}{Task Arithmetic (Top 3 Performances)} \\
\midrule
1st    & 71.4   & 1.2 \\
2nd    & 71.3   & 0.6  \\
3rd    & 70.6   & 0.4  \\
\midrule
\multicolumn{3}{c}{LLM-Designed Algorithms (Top 3 Performances)} \\
\midrule
1st    & \textbf{78.8}  & \textbf{2.5} \\
2nd    & \textbf{78.8}  & \textbf{2.4}  \\
3rd    & \textbf{78.8}  & \textbf{2.0}  \\
\bottomrule
\end{tabular}
}
\caption{Test accuracy (\%) on GSM8k and MATH tasks for Task Arithmetic (top 3 mixing ratios \underline{optimized for new candidate models}) and LLM-discovered algorithms (applying top 15 algorithms from \autoref{sec:main} \underline{without re-optimization for new candidates}).} 
\label{tab:trans_cand_score}
\end{table}

\paragraph{Optimized Task Arithmetic vs. LLM-Discovered Algorithms}
Next, we compare Task Arithmetic that is optimized for the new candidate models, with the LLM-discovered algorithms. 
For Task Arithmetic, we exhaustively explore all combinations of mixing ratios for the new candidate models, following the same procedure as in \autoref{sec:main}.
\autoref{tab:trans_cand_score} provides a detailed comparison of their performance. It is important to note that the algorithms discovered by the LLM are not optimized for the new candidate models (meaning that these models are out-of-domain for these algorithms).

Our algorithms consistently outperform both the individual models and Task Arithmetic across all tasks. In the GSM8k task, our algorithm achieves a high accuracy of 78.8\%, surpassing the best individual model by 3.3 percentage points and the best Task Arithmetic result by 7.4 percentage points. Similarly, in the MATH task, our algorithm reaches 2.5\%, more than doubling the performance of Task Arithmetic. These results not only demonstrate the effectiveness of our proposed method but also highlight its robustness when applied to new model sets without re-optimization. The consistent superiority of our approach over Task Arithmetic, particularly on out-of-domain models, underscores the high performance of the discovered algorithms.

\section{Related Work}
\paragraph{Recursive self-improvement}
The concept of self-improving artificial intelligence was proposed by \citet{Minsky1966} and \citet{DBLP:journals/ac/Good65}, and later formalized by \citet{DBLP:journals/corr/cs-LO-0309048}. With the rapid development of LLMs, the community has shifted towards practical implementations of self-improvement~\cite{DBLP:conf/emnlp/0001GHW00023}. 
Many recent self-improvement approaches primarily focus on the generation of fine-tuning with self-generated training data~\cite{DBLP:conf/icml/YuanPCLSXW24,DBLP:journals/corr/abs-2308-08998,DBLP:journals/corr/abs-2406-03816,DBLP:conf/acl/WangKMLSKH23,DBLP:conf/iclr/XuSZG0FTLJ24}. 
Their methods do not generate or learn the improvement strategies themselves. Additionally, agents that modify outputs based on feedback from the environment have been proposed~\cite{DBLP:conf/nips/MadaanTGHGW0DPY23,DBLP:conf/nips/ShinnCGNY23,DBLP:journals/corr/abs-2404-02183}, but these are different from our goal of improving the LLM itself.
Regarding research on automating LLM development, \citet{yano2025lamdagent} have worked on the automatic construction of post-training pipelines. They employ LLM-based agents to automatically search for combinations of techniques such as SFT and model merging. In contrast, our focus is on the automatic discovery of model improvement algorithms.

\paragraph{Algorithm discovery using LLMs}
Code generation by LLMs~\cite{DBLP:journals/corr/abs-2406-00515} has been proposed for various applications, such as solving reasoning problems~\cite{DBLP:journals/tmlr/ChenM0C23} and generating agent actions~\cite{DBLP:conf/icml/WangCY0L0J24}. 
Code generation has also achieved significant results in scientific discovery.
AlphaEvolve~\cite{NovikovAlphaEvolveA} has demonstrated practical achievements in mathematical discovery and system optimization by combining evolutionary computation with LLMs.
In the context of recursive self-improvement, \citet{DBLP:journals/corr/abs-2406-08414} propose a method for discovering loss functions for preference optimization using LLMs themselves, and \citet{DBLP:journals/corr/abs-2310-02304} propose a method for improving code that makes structured calls to LLMs.
However, existing methods have several important limitations: (1) No generator improvement: The algorithm generator itself is not improved and lacks the ability to learn from generated algorithms to produce better algorithms, (2) External dependency: They rely on more powerful external models such as GPT-4~\cite{DBLP:journals/corr/abs-2303-08774}.
Self-Developing overcomes these limitations by continuously training the algorithm generator (Algorithm Factory) itself using DPO. By using preference data constructed from pairs of high-performing and low-performing algorithms, the algorithm generator learns the characteristics of effective algorithms and becomes capable of efficiently generating higher-quality algorithms over time. Furthermore, it achieves recursive self-improvement without depending on external models other than the seed model.
We are the first to achieve improvement of both the algorithms and the LLM that generates them in the context of recursive self-improvement.
To the best of our knowledge, we are the first to recursively improve an algorithm-generating LLM by training it to generate better algorithms.

\section{Conclusion}
We have proposed \emph{Self-Developing}, a framework for LLMs to autonomously improve through self-generated model-improving algorithms. Our approach does not rely on human expertise or external teacher models. We demonstrated that LLMs can discover superior algorithms that consistently outperform both base models and human-designed algorithms across tasks, and they can apply these algorithms to automatically enhance their own capabilities (\autoref{sec:main}). The algorithms discovered by the LLM also exhibited strong transferability, surpassing both individual models and human-designed algorithms when applied to out-of-domain models (\autoref{sec:trans_result}).
These results suggest that LLMs can independently discover and refine effective model-improving algorithms, paving the way for AI to evolve with minimal human intervention and greater adaptability.

\section*{Limitations}
While our study provides valuable insights, we acknowledge several limitations. First, focusing solely on mathematical reasoning tasks may not fully represent the diverse range of tasks LLMs encounter in real-world applications. Although this choice aligns with standard benchmarks in LLM self-improvement research (e.g., GSM8K and MATH) ~\cite{DBLP:conf/nips/ZelikmanWMG22,DBLP:journals/corr/abs-2406-03816,DBLP:conf/emnlp/0001GHW00023} and allowed for in-depth analysis, extending the evaluation to a broader range of tasks, such as natural language understanding or code generation, could offer additional insights into the generalizability of our findings.
Furthermore, due to computational resource constraints, we had to prioritize specific tasks for in-depth analysis, which prevented us from conducting experiments in other domains. While our results demonstrate LLMs' self-improvement capabilities in mathematical reasoning benchmarks, we recognize the importance of validation across broader domains and hope this gap will be addressed in future research.

\section*{Acknowledgments}
We would like to thank Takuya Tamura and Daichi Haraguchi at NEC Data Science Laboratories for their valuable discussions and insights throughout this research.

\bibliography{custom}

\newpage
\appendix
\section{Transferability}
\label{sec:apdx_trans}
\begin{figure}[H] 
    \centering
    \includegraphics[width=\linewidth]{\thisfolder 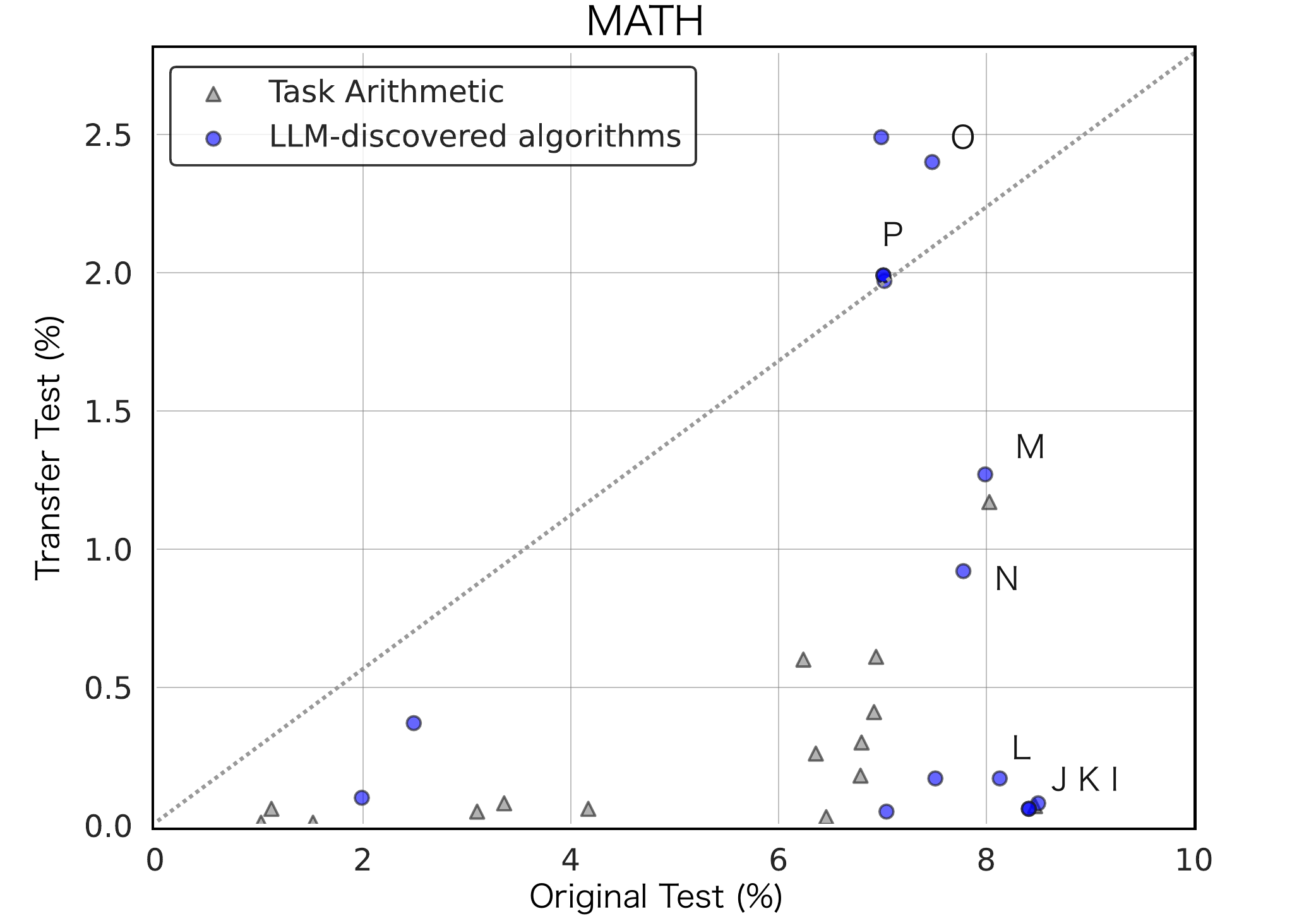} 
    \caption{Transferability of the top 15 merge algorithms for the MATH task. The x-axis shows the test score on original models to be merged, while the y-axis shows the score on new models to be merged. Each point represents a different algorithm, with points above the diagonal line indicating better transferability.}
    \label{fig:transferability_math} 
\end{figure}

Our proposed method demonstrated significant transferability in both the GSM8k and MATH tasks, as shown in \autoref{fig:transferability_gsm8k} and \autoref{fig:transferability_math}. These figures showcase the performance of various discovered algorithms on their original test sets and on transfer test sets with new, unseen merge candidate models. For a more detailed breakdown of algorithm performance, we refer to \autoref{tab:best_algs}.

For the MATH task, most Task Arithmetic scores are below 1\% when applied to new models, indicating the challenge of transferability (\autoref{fig:transferability_math}). In contrast, our generated algorithms achieved scores of up to approximately 2.5\% on new models, significantly outperforming Task Arithmetic.

Transferability for GSM8k is particularly strong. Algorithms A (\autoref{fig:func_example_A}) to D (\autoref{fig:func_example_D}) show improved performance when applied to new models. For instance, Algorithm A (\autoref{fig:func_example_A}) improves from 76.05\% on the original test to 78.75\% on the transfer test. 

A particularly interesting finding is the remarkable cross-task performance of some algorithms. Notably, Algorithm G (\autoref{fig:func_example_G}), discovered using GSM8k data, achieves an accuracy of 74.82\% on GSM8k and 7.96\% on the MATH task. This performance on MATH is nearly on par with Algorithm I (\autoref{fig:func_example_I}), which was specifically optimized for the MATH task (8.50\%). Such cross-task effectiveness suggests the potential for discovering algorithms with LLMs that are effective across various problem types. Additionally, it was found that algorithms discovered for MATH are also effective on GSM8k, suggesting that exploring algorithms on more challenging tasks may lead to the discovery of algorithms that are effective across a broader range of tasks.

\begin{table}[t]
    \centering
    \scalebox{0.8}[0.8]{
    \begin{tabular}{lcccc}
        \toprule
        & \multicolumn{2}{c}{GSM8k} & \multicolumn{2}{c}{MATH} \\
        \cmidrule(lr){2-3} \cmidrule(lr){4-5}
        \makecell{Discovered \\ Algorithms} & \makecell{Original \\Test} & \makecell{Transfer \\Test} & \makecell{Original \\Test} & \makecell{Transfer \\Test} \\
        \midrule
        \midrule
        \multicolumn{5}{c}{Algorithms discovered using GSM8k} \\
        \midrule
        A (\autoref{fig:func_example_A}) & \textbf{76.05} & 78.75 & 2.01 & \textbf{1.72} \\
        B (\autoref{fig:func_example_B}) & 76.05 & 78.75 & 1.99 & 1.63 \\
        C (\autoref{fig:func_example_C}) & 75.96 & 78.75 & 1.82 & 1.67 \\
        D (\autoref{fig:func_example_D}) & 75.96 & \textbf{78.84} & 1.82 & 1.70 \\
        E (\autoref{fig:func_example_E}) & 75.80 & 76.78 & 0.29 & 0.62 \\
        F (\autoref{fig:func_example_F}) & 75.31 & 77.03 & 5.10 & 0.26 \\
        G (\autoref{fig:func_example_G}) & 74.82 & 78.10 & \textbf{7.96} & 0.36 \\
        H (\autoref{fig:func_example_H}) & 74.49 & 74.73 & 6.22 & 1.87 \\
        \midrule
        \multicolumn{5}{c}{Algorithms discovered using MATH} \\
        \midrule
        I (\autoref{fig:func_example_I}) & 69.48 & 69.48 & \textbf{8.50} & 0.08 \\
        J (\autoref{fig:func_example_J}) & 70.30 & \textbf{78.10} & 8.41 & 0.06 \\
        K (\autoref{fig:func_example_K}) & 70.30 & \textbf{78.10} & 8.41 & 0.06 \\
        L (\autoref{fig:func_example_L}) & 69.32 & 63.41 & 8.13 & 0.17 \\
        M (\autoref{fig:func_example_M}) & 69.89 & 53.40 & 7.99 & 1.27 \\
        N (\autoref{fig:func_example_N}) & \textbf{73.83} & 65.14 & 7.78 & 0.92 \\
        O (\autoref{fig:func_example_O}) & 71.29 & 65.87 & 7.48 & \textbf{2.40} \\
        P (\autoref{fig:func_example_P}) & 69.57 & 65.71 & 7.02 & 1.97 \\
        \bottomrule
    \end{tabular}
    }
    \caption{Performance of merged models on GSM8k and MATH tasks. Algorithms A-H were developed using GSM8k data, and algorithms I-P were developed using MATH data. `Original Test' columns show the performance on merge candidate models used in the algorithm search, while `Transfer Test' columns indicate performance on new, unseen merge candidate models, assessing the transferability of each algorithm.}
    \label{tab:best_algs}
\end{table}

\section{Temperature Decay for Iterative DPO}
\label{sec:apdx_temp_decay}

\begin{figure*}[t] 
\centering
\includegraphics[width=11cm]{\thisfolder 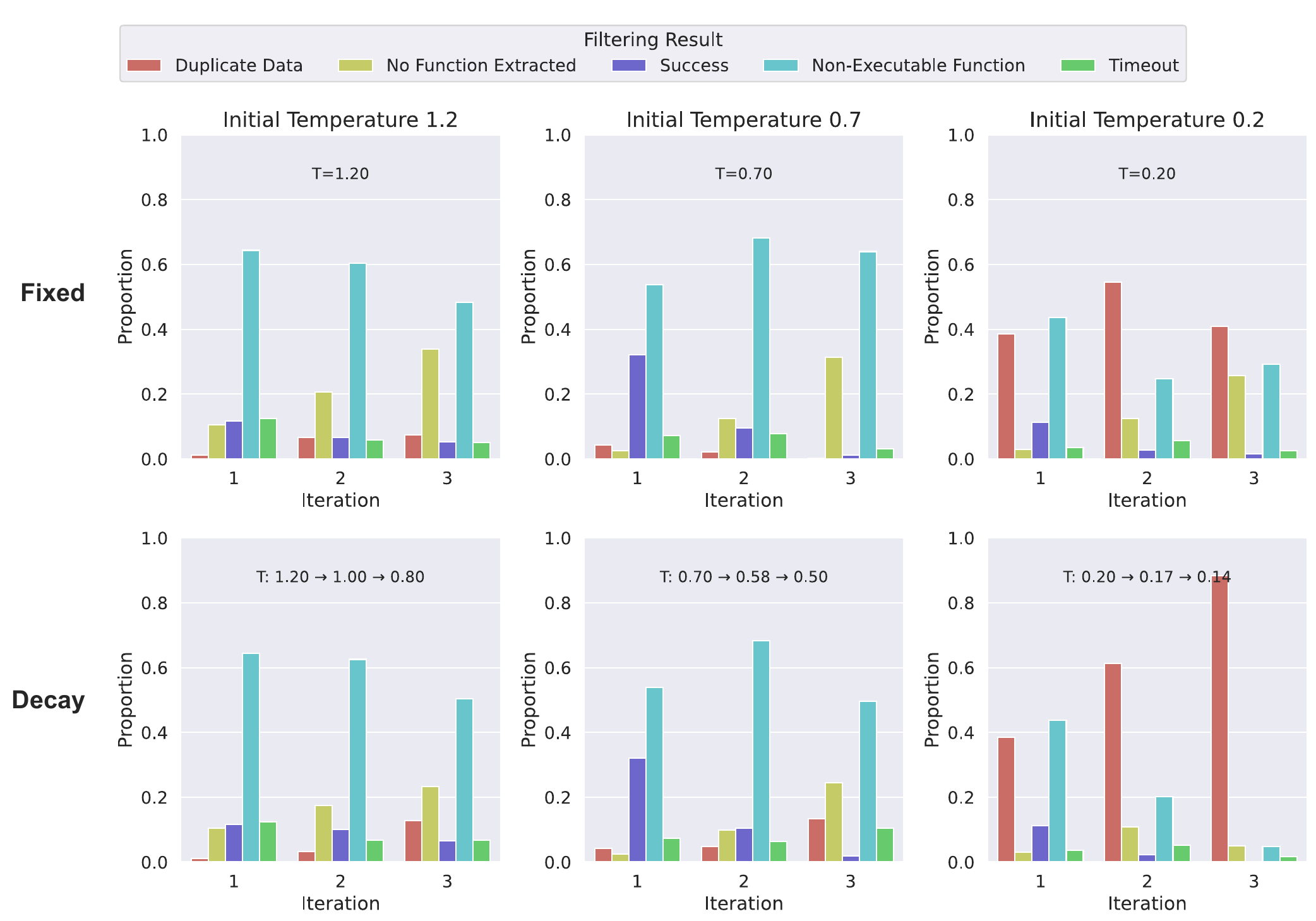}
\caption{Impact of temperature settings and decay on generated Python functions across iterations. Results are categorized as: Duplicate Data, No Function Extracted (failed to generate a function), Success (executable functions), Non-Executable Function (syntactically incorrect), and Timeout (execution time exceeded).}
\label{fig:ablation_temperature_filtering_result}
\end{figure*}

Iterative DPO~\cite{DBLP:conf/icml/YuanPCLSXW24,DBLP:journals/corr/abs-2312-16682,DBLP:conf/icml/YuanPCLSXW24} has been shown to outperform a single round of DPO by iteratively updating the model through preference optimization steps, thus producing refined outputs.

Temperature~\cite{DBLP:journals/corr/HintonVD15} is a crucial parameter for controlling the creativity of the generated text. In our method, it also plays a significant role in the generation process of model-improving algorithms.
Generally, a higher temperature in LLMs results in more diverse and creative text, while a lower temperature yields more accurate outputs~\cite{openai2023prompt}. This can be viewed as a means to control the trade-off between \emph{exploration} and \emph{exploitation}.
Recent studies have proposed methods to dynamically adjust the temperature based on the input text~\cite{DBLP:conf/aaai/ZhuLLZLJ024,DBLP:conf/aaai/JoyPLTD23,DBLP:journals/corr/abs-2012-13575}. 
In iterative DPO, the temperature has traditionally been set manually\footnote{For instance, in \cite{DBLP:conf/icml/YuanPCLSXW24}, the temperature is fixed at $T = 0.6$ or $T = 0.7$ during data generation step for iterative DPO.}.

To appropriately balance exploration and exploitation during the algorithm generation process, we introduce a temperature decay  inspired by learning rate decay~\cite{DBLP:conf/nips/0001KKN19, DBLP:conf/iclr/LoshchilovH17}. 
This approach allows for dynamic adjustment of the exploration strategy as iterations progress. 

In the initial iterations, a high initial temperature facilitates the generation of a wide range of creative algorithms, maximizing the opportunity to discover innovative solutions that might be overlooked by conventional approaches. During the mid-phase, a gradual decrease in temperature leverages the effective features of the algorithms learned so far while continuing to explore new variations and combinations. In the later stages, a lower temperature focuses the search around known high-performance algorithms, increasing the likelihood of efficiently discovering superior algorithms.

Specifically, the temperature update at iteration $t$ is based on the Inverse Time Decay schedule:
\begin{equation}
    T_t = \frac{T_1}{1 + \beta (t - 1)},
\end{equation}
where $T_1$ denotes the initial temperature, and $\beta \in \mathbb{R}$ is a hyperparameter that controls the decay rate. By adjusting the decay rate $\beta$, one can regulate the speed of the transition from exploration to exploitation.

\paragraph{Experiment}
This experiment investigates the impact of temperature settings and their decay on the quality and diversity of Python functions generated in an iterative DPO process. 
\autoref{fig:ablation_temperature_filtering_result} visualizes the filtering results to observe qualitative changes in Python functions sampled from the algorithm factory $\pi_t^g$ at each iteration.
The figure shows the results for different temperature settings with and without temperature decay.
The experiment was conducted under the following conditions:
\begin{itemize}
\item Initial temperatures: $\{1.20, 0.70, 0.20\}$
\item Temperature control: Fixed temperature and temperature decay ($\beta = 0.2$)
\item Number of iterations: 3 for each condition
\end{itemize}

The generated Python functions were filtered into the following categories:
\begin{itemize}
\item Duplicate Data
\item No Function Extracted
\item Success (executable functions)
\item Non-Executable Function
\item Timeout
\end{itemize}

\paragraph{Key finding 1: Higher temperatures is effective for enhancing data diversity}
Comparing high and low temperature settings, it was found that higher temperatures consistently produce more diverse data. Throughout all iterations, low temperature ($T=0.20$) tends to generate a higher proportion of duplicate data, reducing diversity. In contrast, high temperatures ($T=0.70$, $T=1.20$) produce less duplication and more diverse data.
The $T=0.70$ setting generates the highest number of executable functions (`Success' in \autoref{fig:ablation_temperature_filtering_result}) in the first iteration, but this proportion decreases sharply in later iterations. The $T=1.20$ setting, while having a lower initial success rate, continues to generate a relatively high number of executable functions in later iterations. These results suggest that higher temperature settings can generate high-quality data more consistently over the long term.

\paragraph{Key finding 2: Temperature decay is effective for stable data generation}
Applying temperature decay tends to be more effective than using a fixed temperature for generating data stably. With fixed temperatures, there is a tendency for the rate of non-executable functions to increase in later iterations. When temperature decay is applied, the rate of duplicate functions shows an increase in later iterations, but the rate of non-executable functions decreases, resulting in a small increase in the number of executable algorithms ('Success').
This phenomenon suggests that temperature decay may shift the generation process from creating more varied data towards generating  more accurate data. These findings indicate that an appropriate temperature decay strategy could play a role in optimizing the quality and diversity of generated data in iterative DPO.

\section{Analysis of LLM-Discovered Algorithms}
\label{sec:alg_analysis}
\paragraph{Complexity of Coefficient Calculation}
In model merging, coefficients play a crucial role in determining how different models are merged. 
The coefficients in merge strategies mainly included the following: (1) Weighting Factor, determining the extent to which weights of different models are reflected, (2) Adaptive Coefficient, dynamically adjusted based on model characteristics (e.g., weight norms), and (3) Blend Ratio, determining the ratio when combining different merge strategies (e.g., multiplication and averaging).
For example:
\begin{figure}[H]
  \centering
\begin{mybox}[]
\lstset{basicstyle=\scriptsize\ttfamily}
\begin{lstlisting}
# Weighting Factor
x + alpha * (y - x)

# Adaptive Coefficient
alpha * torch.mul(x, y) + beta * torch.max(x, y)

# Blend Ratio
(average + element_wise_maximum + element_wise_minimum) / alpha
\end{lstlisting}
\end{mybox}
\end{figure}
There was a marked tendency for coefficient calculations to become more complex as iterations progressed. In iteration $t=1$, relatively simple coefficients (e.g., a fixed value of 0.5) were often used for mixing task vectors, but by iteration $t = 2$, a method was introduced for dynamically calculating coefficients using the cosine similarity of task vectors (Algorithm O; \autoref{fig:func_example_O}), similar to Model Stock~\cite{DBLP:conf/eccv/JangYH24}.
The increasing complexity of coefficient calculations may enable the realization of more precise and adaptive merge strategies. This likely resulted in a tendency to enhance performance by fine-tuning coefficients while maintaining the basic structure of specific strategies.

\paragraph{Diversity of Ideas}
In the early iterations, a wide range of ideas were explored. \autoref{tab:strategy_words} shows a portion of the words and their frequencies found in the strategies of algorithms generated by the LLM during iteration $t=1$. This result demonstrates that diverse methods are proposed. The most frequently used methods are based on 'similarity' and 'distance'. There is a clear tendency to utilize geometric information of vectors ('angle', 'geometric', 'metric', 'norm', 'frobenius', etc.). 
Additionally, 'element-wise' and 'pairwise' operations are also commonly observed. Furthermore, a wide array of algorithms are proposed, including statistical methods ('kullback', 'leibler', 'gaussian', 'distribution', 'entropy', 'lasso', etc.), learning-based approaches ('learning', 'train'), matrix decomposition ('factorization', 'svd', 'pca'), and grouping techniques ('clustering',  'neighbor', 'kmeans', etc.).
Among the creative algorithms, many interesting ones are included. For example, the set similarity-based method is a unique algorithm that treats vectors as sets of values and calculates their set similarity (\autoref{fig:func_example_unique1}). Although the development scores of models using these methods are not high, there is potential to discover superior algorithms by increasing the number of generated algorithms.

\begin{table*}[ht]
\centering
    \scalebox{0.8}[0.8]{
\begin{tabular}{lc|lc|lc|lc}
\toprule
\textbf{Word} & \textbf{Frequency} & \textbf{Word} & \textbf{Frequency} & \textbf{Word} & \textbf{Frequency} & \textbf{Word} & \textbf{Frequency} \\
\midrule
weight           & 564 & group          & 10 & probability    & 4 & pooling        & 2 \\
similarity       & 285 & attention      & 10 & sequence       & 4 & softmax        & 2 \\
distance         & 262 & variance       & 10 & correlation    & 4 & dropout        & 2 \\
mean             & 217 & factorization  & 9  & absolute       & 4 & euclidean      & 2 \\
norm             & 158 & metric         & 9  & pca            & 4 & intersection   & 2 \\
average          & 61  & learning       & 9  & clustering     & 4 & zscore         & 1 \\
element          & 40  & decomposition  & 8  & frobenius      & 3 & ode            & 1 \\
maximum          & 38  & decay          & 8  & voting         & 3 & moment         & 1 \\
l1               & 32  & magnitude      & 8  & lp             & 3 & tikhonov       & 1 \\
sum              & 27  & median         & 8  & regression     & 3 & lasso          & 1 \\
minimum          & 26  & domain         & 7  & neighbor       & 3 & ridge          & 1 \\
wise             & 23  & hybrid         & 7  & gradient       & 3 & polymorphism   & 1 \\
difference       & 22  & pairwise       & 7  & train          & 3 & skewness       & 1 \\
matrix           & 19  & entropy        & 6  & kernel         & 3 & kurtosis       & 1 \\
normalization    & 16  & means          & 6  & hadamard       & 3 & guessscore     & 1 \\
cluster          & 16  & distribution   & 6  & ema            & 3 & sigmoid        & 1 \\
optimization     & 14  & kl             & 5  & tucker         & 2 & ghash          & 1 \\
dimension        & 13  & heuristic      & 5  & leibler        & 2 & newton         & 1 \\
coefficient      & 13  & order          & 5  & kullback       & 2 & svd            & 1 \\
scale            & 11  & geometric      & 5  & trimean        & 2 & sort           & 1 \\
addition         & 10  & angle          & 5  & approximation  & 2 & rmse           & 1 \\
threshold        & 10  & rank           & 4  & tree           & 2 & pivot          & 1 \\
regularization   & 10  & moving         & 4  & hamming        & 2 & noise          & 1 \\
\bottomrule
\end{tabular}
}
\caption{Word frequency in comments of Python code generated by the algorithm factory at iteration $t=1$. These words (nouns) were extracted from comments following the prefix (\texttt{\# New strategies for merging the model weights:}).}
\label{tab:strategy_words}
\end{table*}

\begin{figure*}[t]
  \centering
\vspace{-10pt}
\begin{mybox}[Discovered Merging Function]
\vspace{-9pt}
\lstset{basicstyle=\scriptsize\ttfamily}
\begin{lstlisting}
def merge_models(model_dict, device):
    ''' Develop and implement a novel algorithm...(omitted)'''
    # *New* strategies for merging the model weights:
    # 1. Consider calculating the norms (sizes) of the weight tensors.
    # 2. Define a weighting function that takes into account both model consistency and diversity.
    # 3. Introduce a parameter `p`, adjusting the balance between model consistency and diversity.
    # 4. Introduce another parameter `alpha` adjusting the balance between linear interpolation and weighted averaging.

    # Assign parameters `p` and `alpha`
    p = 0.75
    alpha = 0.5
    
    # Convert model weights to tensors and move them to the specified device (CPU or GPU)
    weights = [model.detach().to(device) for model in model_dict.values()]
    
    def weighting_function(x, y, p, alpha):
        # Calculate weight using the Jaccard similarity
        intersection = torch.sum(torch.sigmoid(x) * torch.sigmoid(y))
        union = torch.sum(torch.sigmoid(x)) + torch.sum(torch.sigmoid(y))
        jaccard = intersection / union
        # Normalize the weights using weighting parameter `p`
        normalized_jaccard = jaccard ** p
        # Combine weights using a mix of interpolation and averaging with parameter `alpha`
        return alpha * x + (1 - alpha) * normalized_jaccard * y
        
    # Initialize merged_weights with the first model's weights
    merged_weights = weights[0].clone()
    for weight in weights[1:]:
        merged_weights = weighting_function(merged_weights, weight, p, alpha)
    
    return merged_weights
\end{lstlisting}
\vspace{-6pt}
\end{mybox}
\vspace{-9pt}
  \caption{This algorithm demonstrates a creative approach. The vectors are interpreted as sets of values, with the Jaccard index serving as a similarity measure for adaptive weighting.}
  \label{fig:func_example_unique1}
\end{figure*}

\paragraph{Merging strategy: Algorithm A}  
We explain the merging algorithm that achieved the best performance on the GSM8k task among the generated algorithms, demonstrating exceptionally high transferability to out-of-domain models (labeled `A' in \autoref{fig:transferability_gsm8k}).  
The exact function generated can be found in \autoref{fig:func_example_A}.  
Below, we present a mathematical formulation of the algorithm.

This merging algorithm repeatedly applies a function (implemented as \texttt{hybrid\_merge\_strategy} in the code) to sequentially merge the task vectors. Starting with the initial vector $\tau_1^{{\text{merged}}} = \tau_1$, the function $f$ is applied iteratively to combine each subsequent task vector $\tau_i$ with the current merged vector. This process can be represented as follows:
\begin{equation}
\begin{aligned}
\tau_2^{{\text{merged}}} &= f(\tau_1^{{\text{merged}}}, \tau_2), \\
\tau_3^{{\text{merged}}} &= f(\tau_2^{{\text{merged}}}, \tau_3), \\
&\vdots \\
\tau_K^{{\text{merged}}} &= f(\tau_{K-1}^{{\text{merged}}}, \tau_K).
\end{aligned}
\end{equation}
Finally, this algorithm outputs the final merged vector $\tau_K^{{\text{merged}}}$.
Here, the function $f$ can be defined as: 
\begin{equation}
f(\tau_{i-1}^{{\text{merged}}}, \tau_i) = \frac{1}{2}\left( \tau_{i-1}^{{\text{merged}}} + \mu_i \mathbf{1} \right), 
\end{equation}
where $d$ is the dimension of the task vectors, and $\mathbf{1} \in \mathbb{R}^d$ is a vector with all elements are 1.
$\mu_i \in \mathbb{R}$ is the mean of all elements of the task vector $\tau_i$: 
\begin{equation}
    \mu_i = \frac{1}{d} \sum_{j=1}^{d} (\tau_i)_j,
\end{equation}
where $(\tau_i)_j \in \mathbb{R}$ denotes the $j$-th element of $\tau_i$.

\begin{figure*}[t]
  \centering
\vspace{-10pt}
\begin{mybox}[Discovered algorithm A]
\vspace{-9pt}
\lstset{basicstyle=\scriptsize\ttfamily}
\begin{lstlisting}[language=Python]
def merge_models(model_dict, device):
    ''' Develop and implement a novel algorithm...(omitted)'''
    # *New* strategies for merging the model weights:
    # 1. Hybrid approach using element-wise multiplication and average
    
    # Convert model weights to tensors and move them to the specified device (CPU or GPU)
    weights = [model.detach().to(device) for model in model_dict.values()]

    def hybrid_merge_strategy(x, y, alpha=0.5):
        # Calculate element-wise multiplication and average
        return (1 - alpha) * x + alpha * torch.mean(y, dim=(0 if x.dim() == 1 else 1), keepdim=True)

    # Iteratively apply the merge strategy to combine each subsequent model's weights with the initial model's weights
    initial_weights = weights[0].clone()
    merged_weights = weights[0].clone()
    
    for i in range(len(weights)):
        if i == 0:
            continue
        merged_weights = hybrid_merge_strategy(merged_weights, weights[i], alpha=0.5)
        # Store the merged weights after every k-th model
        if i % len(weights) == 0:
            weights[0] = merged_weights.clone()

    return merged_weights
\end{lstlisting}
\vspace{-6pt}
\end{mybox}
\vspace{-9pt}
\caption{Discovered algorithm A. This is one of the most effective algorithms discovered by the LLM, generated during iteration $t=3$.}
\label{fig:func_example_A}
\end{figure*}

\clearpage
\section{Pseudocode}
\label{sec:pseudocode}
\begin{algorithm*}
\small
\begin{algorithmic}[1]
    \Require $M_0$: Seed model, $\mathcal{T}$: Target Task
    \Require $x$: Prompt
    \Require $I$: Max iterations, $N$: Number of algorithm generation
    \Require $T_1$: Initial temperature, $\beta$: Decay rate
    \Require $p_w$, $p_l$: Percentages for DPO data selection
    \Require $k$: Number of top-performing algorithms to add from previous iterations
    \Require $S$: Number of low-performing algorithms to pair with each high-performing algorithm
    \Ensure $M^{{\text{best}}}$: Best improved model
    \State Initialize algorithm generator: $\pi_1^g \gets M_0$
    \State Initialize best model: $M^{{\text{best}}} \gets M_0$
    \State Initialize best score: $s_{\text{best}} \gets -\infty $
    
    \For{$t = 1$ to $I$}
        \State \textcolor[HTML]{4C4CFF}{\texttt{// Algorithm Generation}}
        \State Update temperature: $T_t = \frac{T_1}{1 + \beta (t - 1)}$  \Comment{Decrease temperature}
        \State $A_t \gets \{\}$  \Comment{Initialize empty set for algorithms}
        \For{$i = 1$ to $N$}
            \State $a_t^{(i)} \sim \pi_t^g(a \mid x)$  \Comment{Generate algorithm with temperature $T_t$}
            \If{\text{IsValid}($a_t^{(i)}$)}
                \State $A_t \gets A_t \cup \{a_t^{(i)}\}$  \Comment{Add valid algorithm to set}
            \EndIf
        \EndFor
        
        \State \textcolor[HTML]{4C4CFF}{\texttt{// Algorithm Evaluation}}
        \State $S_t \gets \{\}$  \Comment{Initialize empty set for scores}
        \For{$a_t^{(i)} \in A_t$}
            \State $M_t^{(i)} \gets \text{Apply}(a_t^{(i)}, M_0)$  \Comment{Apply algorithm to get improved model}
            \State $s_t^{(i)} \gets \text{Evaluate}_{\mathcal{T}}(M_t^{(i)})$  \Comment{Evaluate improved model with dev set}
            \State $S_t \gets S_t \cup \{s_t^{(i)}\}$  \Comment{Add task score to set}
        \EndFor

        \State \textcolor[HTML]{4C4CFF}{\texttt{// DPO Data Selection}}
        \State $s_{p_w} \gets \text{Percentile}(S_t, 100-p_w)$  \Comment{Top $p_w\%$ score threshold}
        \State $s_{p_l} \gets \text{Percentile}(S_t, p_l)$  \Comment{Bottom $p_l\%$ score threshold}
        \State $A_{t,w} \gets \{a_t^{(i)} \in A_t \mid s_t^{(i)} \geq s_{p_w}\}$  \Comment{High-performing algorithms}
        \State $A_{t,l} \gets \{a_t^{(i)} \in A_t \mid s_t^{(i)} \leq s_{p_l}\}$  \Comment{Low-performing algorithms}
        
        \State $A_{\text{pre}, w} \gets \bigcup_{j=1}^{t-1} A_{j,w}$  \Comment{Union of all previous high-performing algorithms}
        \State $A_{\text{top3}} \gets \text{TopK}(A_{\text{pre},w}, k)$  \Comment{Select top 3 algorithms based on scores}
        \State $A_{t,w} \gets A_{t,w} \cup A_{\text{top3}}$  \Comment{Add top 3 to high-performing set}
        
        \State $\mathcal{D} \gets \{\}$  \Comment{Initialize empty DPO dataset}
        \For{$a_{t,w}^{(i)} \in A_{t,w}$}
            \State $L_i \gets \text{Sample}(A_{t,l}, S)$  \Comment{Sample $S$ low-performing algorithms}
            \For{$a_{t,l}^{(j)} \in L_i$}
                \State $\mathcal{D} \gets \mathcal{D} \cup \{(x, a_{t,w}^{(i)}, a_{t,l}^{(j)})\}$  \Comment{Add pair to DPO dataset}
            \EndFor
        \EndFor
        
        \State \textcolor[HTML]{4C4CFF}{\texttt{// Update Algorithm Generator}}
        \State Update $\pi_t^g$ to $\pi_{t+1}^g$ using DPO with $\mathcal{D}$
        
        \State \textcolor[HTML]{4C4CFF}{\texttt{// Update Best Model}}
        \If{$\max(S_t) > s_{\text{best}}$}
            \State $s_{\text{best}} \gets \max(S_t)$
            \State $M^{\text{best}} \gets \text{Apply}(a_t^{(i^*)}, M_0)$ where $i^* = \arg\max_i s_t^{(i)}$
        \EndIf
    \EndFor
    \State \Return $M^{{\text{best}}}$
\end{algorithmic}
\caption{Self-Developing}
\label{alg:framework}
\end{algorithm*}

\clearpage
\section{Prompt}\label{sec:prompt}
\autoref{fig:prompt_template} shows the prompt template we used for generating model merging algorithms. The prompt uses a relatively creative merging algorithm as a one-shot example (\autoref{fig:one_hot_example}). While simpler examples might seem sufficient, our preliminary experiments suggested the need for a more sophisticated example to guide the generation of creative merging algorithms.

During our preliminary experiments, we investigated how the presence or absence of a one-shot example affects algorithm generation. This example serves multiple purposes: demonstrating the expected format of a Python function, showing how to handle model weights as tensors, and illustrating basic weight combination operations.

Our preliminary exploration of zero-shot settings (i.e., without the one-shot example) revealed several important challenges. Many generated outputs failed to be executable Python functions, often containing syntax errors or undefined variables. The generated algorithms also showed less variety in their approaches, mostly converging to simple weighted averaging operations.

These preliminary findings led us to include the one-shot example in our main experiments, as it appeared crucial not only for ensuring the generation of executable code but also for encouraging the exploration of diverse algorithmic strategies. The example helps the LLM understand both the technical requirements (e.g., proper tensor operations) and the potential space of solutions for model merging algorithms.

\begin{figure*}[t]
  \centering
\begin{mybox}[One-shot example]
\lstset{basicstyle=\scriptsize\ttfamily}
\begin{lstlisting}[numbers=none, xleftmargin=0pt, framexleftmargin=0pt]
def merge_models(model_dict, device):
    '''
    Develop and implement a novel algorithm for merging the model weights. Your goal is to create a unique and effective strategy that combines various existing techniques and introduces new approaches to achieve optimal performance. Consider integrating methods such as adaptive weighting, hybrid strategies, or advanced heuristics to create a more innovative merging technique.

    Args:
        model_dict (dict): A dictionary where keys are model names and values are the model weights.
        device (torch.device): The device (CPU or GPU) on which the computation will be performed.
        
    Returns:
        torch.Tensor: The weights of the merged model.
    '''
    # Strategy for merging the model weights:
    # 1. Initialize `merged_weights` with the first model's weights.
    # 2. Iteratively apply the merge strategy to combine each subsequent model's weights with the merged result.
    
    # Convert model weights to tensors and move them to the specified device (CPU or GPU)
    weights = [model.detach().to(device) for model in model_dict.values()]

    def merge_strategy(x, y):
        # Calculate the norms (sizes) of the weight tensors
        x_size = torch.norm(x)
        y_size = torch.norm(y)
        # Adjust the weighting factor based on the norms
        alpha = (x_size + y_size) * 0.5 / x_size
        # Merge the weights using the adjusted alpha
        return x + alpha * y

    # Initialize merged_weights with the first model's weights
    merged_weights = weights[0].clone()
    
    # Iteratively merge each subsequent model's weights
    for weight in weights[1:]:
        merged_weights = merge_strategy(merged_weights, weight)

    return merged_weights
\end{lstlisting}
\end{mybox}
\caption{One-shot example.}
\label{fig:one_hot_example}
\end{figure*}

\begin{figure*}[t]
  \centering
\begin{mybox}[Prompt Template]
\# Task\par
The goal is to merge the weights of multiple pre-trained language models to create a merged model that effectively combines the weights of different models to achieve higher performance. Refer to the code below and devise a new merging strategy to implement.

\mbox{}
\mbox{}

\#\# Reference Code\par
```python
\lstset{basicstyle=\scriptsize\ttfamily}
\begin{lstlisting}[numbers=none, xleftmargin=0pt, framexleftmargin=0pt, belowskip=-1em]
import torch

models = {
    'GAIR/Abel-7B-002': torch.rand(dim), # Abel-7B-002 is a model fine-tuned for mathematical tasks, demonstrating strong performance on datasets such as GSM8k and MATH.
    'SciPhi/SciPhi-Mistral-7B-32k': torch.rand(dim), # SciPhi-Mistral-7B-32k is a fine-tuned LLM focused on scientific reasoning and education, optimized for Alpaca-style prompts.
    'teknium/OpenHermes-2.5-Mistral-7B': torch.rand(dim), # OpenHermes 2.5 is a fine-tuned model, building on OpenHermes 2, specifically enhanced with additional code datasets. Training on code improved its performance on various non-code benchmarks like TruthfulQA and AGIEval.
}

def merge_models(model_dict, device):
    '''
    Args:
        model_dict (dict): A dictionary where keys are model names and values are the model weights.
        device (torch.device): The device (CPU or GPU) on which the computation will be performed.

    Returns:
        torch.Tensor: The weights of the merged model.
    '''
    # Implement the merging strategy here
\end{lstlisting}
```

\#\# Implementation Instructions\par
Implement the `merge\_models` function and devise a new strategy for merging the model weights. Consider combining multiple strategies such as weighted averages, element-wise maximums, element-wise minimums, geometric means, Manhattan distances (L1 norm), cosine similarity, Euclidean distances (L2 norm), harmonic means, median merging, matrix factorization, or hadamard product. Document your thought process and the changes you make in the code.

\mbox{}

\#\#\# Example1\par
```python\par
\{One-shot exaple\}\par
```

\mbox{}

\#\#\# Example2\par
```python
\lstset{basicstyle=\scriptsize\ttfamily}
\begin{lstlisting}[numbers=none, xleftmargin=0pt, framexleftmargin=0pt]
def merge_models(model_dict, device):
    '''
    Develop and implement a novel algorithm for merging the model weights. Your goal is to create a unique and effective strategy that combines various existing techniques and introduces new approaches to achieve optimal performance. Consider integrating methods such as adaptive weighting, hybrid strategies, or advanced heuristics to create a more innovative merging technique.

    Args:
        model_dict (dict): A dictionary where keys are model names and values are the model weights.
        device (torch.device): The device (CPU or GPU) on which the computation will be performed.
        
    Returns:
        torch.Tensor: The weights of the merged model.
    '''
    # *New* strategies for merging the model weights:
\end{lstlisting}
\end{mybox}
\caption{Prompt template.}
\label{fig:prompt_template}
\end{figure*}

\clearpage


\section{Discovered Algorithms}
\label{sec:apdx_algorithms}

\begin{figure*}[t]
  \centering
\vspace{-10pt}
\begin{mybox}[Discovered algorithm B]
\vspace{-9pt}
\lstset{basicstyle=\scriptsize\ttfamily}
\begin{lstlisting}[language=Python]
def merge_models(model_dict, device):
    ''' Develop and implement a novel algorithm...(omitted)'''
    # *New* strategies for merging the model weights:
    # 1. Hybrid approach combining element-wise multiplication and average
    
    # Convert model weights to tensors and move them to the specified device (CPU or GPU)
    weights = [model.detach().to(device) for model in model_dict.values()]

    def hybrid_merge_strategy(base, to_merge, alpha=0.5):
        average = (base + torch.mean(to_merge, dim=0) * alpha) / (1 + alpha)
        weighted = torch.mean(to_merge * torch.tensor(1 / alpha, device=base.device).unsqueeze(0), dim=0)
        return (1 - alpha) * base + alpha * weighted * torch.tensor(alpha, device=base.device).unsqueeze(0)

    # Iteratively merge the weights using the custom strategy
    merged_weights = weights[0].clone()
    for i in range(1, len(weights)):
        merged_weights = hybrid_merge_strategy(merged_weights, weights[i])

    return merged_weights

\end{lstlisting}
\vspace{-6pt}
\end{mybox}
\vspace{-9pt}
\caption{Discovered algorithm B. }
\label{fig:func_example_B}
\end{figure*}

\begin{figure*}[t]
  \centering
\vspace{-10pt}
\begin{mybox}[Discovered algorithm C]
\vspace{-9pt}
\lstset{basicstyle=\scriptsize\ttfamily}
\begin{lstlisting}[language=Python]
def merge_models(model_dict, device):
    ''' Develop and implement a novel algorithm...(omitted)'''
    # *New* strategies for merging the model weights:
    # - Define a merge strategy using a hybrid approach that incorporates element-wise multiplication and weighted averaging
    # - Introduce an additional parameter `alpha` that can be tuned to control the contribution of each constituent model
    # - Utilize a validation dataset to dynamically adjust `alpha` based on the performance improvement on the validation set
    
    # Extract weights from the models and move them to the specified device
    weights = [model.detach().to(device) for model in model_dict.values()]

    def merge_strategy(x, y, alpha):
        # Apply element-wise multiplication
        product = torch.mul(x, y)
        # Perform weighted averaging
        return torch.mul(product, alpha) + torch.mul(1 - alpha, x)

    # Define a function to evaluate the performance of the merged model on a validation set
    def validate_model(model, valid_dataloader):
        # Implement the validation logic
        pass

    # Initialize `alpha` with a default value (e.g., 0.5) or a value obtained from a preliminary experiment
    alpha = 0.5
    # Alternatively, `alpha` can be dynamically adjusted using a validation dataset
    
    # Initialize merged_weights with the first model's weights
    merged_weights = weights[0].clone()
    
    # Iteratively merge each subsequent model's weights using the new hybrid strategy
    for i, weight in enumerate(weights[1:], 1):
        # Adjust `alpha` based on the performance improvement (optional)
        # new_alpha = adaptive_alpha_tuning(alpha, validate_model(model, valid_dataloader), model_dict.keys()[i])
        # merged_weights = merge_strategy(merged_weights, weight, new_alpha)
        
        # Merge the weights using the hybrid strategy with the current alpha value
        merged_weights = merge_strategy(merged_weights, weight, alpha)

    return merged_weights

\end{lstlisting}
\vspace{-6pt}
\end{mybox}
\vspace{-9pt}
\caption{Discovered algorithm C. }
\label{fig:func_example_C}
\end{figure*}

\begin{figure*}[t]
  \centering
\vspace{-10pt}
\begin{mybox}[Discovered algorithm D]
\vspace{-9pt}
\lstset{basicstyle=\scriptsize\ttfamily}
\begin{lstlisting}[language=Python]
def merge_models(model_dict, device):
    ''' Develop and implement a novel algorithm...(omitted)'''
    # *New* strategies for merging the model weights:
    # 1. Design a hybrid strategy by performing element-wise multiplication and mean
    # 2. Define two parameters, alpha and beta, to control the merging ratio
    
    # Convert model weights to tensors and move them to the specified device (CPU or GPU)
    weights = [model.detach().to(device) for model in model_dict.values()]
    
    # Define two parameters to control the merging ratio
    alpha = 0.6
    beta = 0.4
    
    def merge_strategy(x, y, alpha=0.5, beta=0.5):
        # Perform element-wise multiplication
        xy = x * y
        # Perform mean aggregation to find the average weights
        return alpha * x + beta * torch.mean(xy, dim=0)

    # Initialize merged_weights with the first model's weights
    merged_weights = weights[0].clone()
    
    # Iteratively merge each subsequent model's weights
    for weight in weights[1:]:
        merged_weights = merge_strategy(merged_weights, weight)
    
    return merged_weights

\end{lstlisting}
\vspace{-6pt}
\end{mybox}
\vspace{-9pt}
\caption{Discovered algorithm D. }
\label{fig:func_example_D}
\end{figure*}

\begin{figure*}[t]
  \centering
\vspace{-10pt}
\begin{mybox}[Discovered algorithm E]
\vspace{-9pt}
\lstset{basicstyle=\scriptsize\ttfamily}
\begin{lstlisting}[language=Python]
def merge_models(model_dict, device):
    ''' Develop and implement a novel algorithm...(omitted)'''
    # *New* strategies for merging the model weights:
    # - Calculate cosine similarity (allow vector embedding models with different dimensions)
    # - Perform Harmonic mean (in scenarios where average performs poorly due to rare peaks)

    def cosine_similarity(v1, v2):
        return (v1 * v2).sum() / ((v1 ** 2).sum() * (v2 ** 2).sum()) ** 0.5

    def harmonic_mean(y_pred, labels):
        y_pred = torch.clamp(y_pred, 1e-5, 1.0)  # avoid zero division
        return (labels.size(0) + (labels * y_pred).sum()).float() / (labels.sum() + y_pred.sum())
    
    weights = [model.detach().to(device) for model in model_dict.values()]

    # Start merging from the second weight.
    for i in range(1, len(weights)):
        weight = weights[i]
        last_weight = weights[i-1]

        # Calculate the cosine similarity as the merging strategy
        sim = cosine_similarity(weight.reshape(-1), last_weight.reshape(-1))
        
        # Perform element-wise multiplication according to the similarity
        last_weight *= sim
        
        # Save for next merge
        weights[i] = last_weight

    # Last merged weights
    merged_weights = weights[-1]
    return merged_weights

\end{lstlisting}
\vspace{-6pt}
\end{mybox}
\vspace{-9pt}
\caption{Discovered algorithm E. }
\label{fig:func_example_E}
\end{figure*}

\begin{figure*}[t]
  \centering
\vspace{-10pt}
\begin{mybox}[Discovered algorithm F]
\vspace{-9pt}
\lstset{basicstyle=\scriptsize\ttfamily}
\begin{lstlisting}[language=Python]
def merge_models(model_dict, device):
    ''' Develop and implement a novel algorithm...(omitted)'''
    # *New* strategies for merging the model weights:
    # - Hybrid approach: combine element-wise multiplication with average
    
    # Convert model weights to tensors and move them to the specified device (CPU or GPU)
    weights = [model.detach().to(device) for model in model_dict.values()]

    def hybrid_merge_strategy(base, to_merge, alpha=0.5):
        # Calculate the average weights
        average = (base + torch.mean(to_merge, dim=0) * 0.5)
        # Scale the weights to be added but keep the important weights from the top models larger
        return base + alpha * (torch.mean(to_merge, dim=0) - base) * 0.5

    merged_weights = weights[0].clone()

    # Sort the weights based on the norms (sizes) of the weight tensors in descending order
    weights.sort(key=lambda x: torch.norm(x), reverse=True)

    # Iteratively merge the weights with the current merged_weights
    for weight in weights:
        merged_weights = hybrid_merge_strategy(merged_weights, weight)

    return merged_weights

\end{lstlisting}
\vspace{-6pt}
\end{mybox}
\vspace{-9pt}
\caption{Discovered algorithm F. }
\label{fig:func_example_F}
\end{figure*}

\begin{figure*}[t]
  \centering
\vspace{-10pt}
\begin{mybox}[Discovered algorithm G]
\vspace{-9pt}
\lstset{basicstyle=\scriptsize\ttfamily}
\begin{lstlisting}[language=Python]
def merge_models(model_dict, device):
    ''' Develop and implement a novel algorithm...(omitted)'''
    # *New* strategies for merging the model weights:
    # - Implement a hybrid strategy that combines multiple methods, such as element-wise multiplication and averaging
    
    # Convert model weights to tensors and move them to the specified device (CPU or GPU)
    weights = [model.detach().to(device) for model in model_dict.values()]

    def hybrid_strategy(x, y, alpha):
        # Perform element-wise multiplication
        z = torch.mul(x, y)
        # Perform averaging
        z_avg = (x + y) * 0.5
        # Adjust weights with factor alpha
        return z * alpha + z_avg * (1 - alpha)

    # Define a function to calculate weighting factors based on weight tensor norms
    def calculate_alpha(x, y):
        x_size = torch.norm(x)
        y_size = torch.norm(y)
        return (x_size + y_size) * 0.5 / (x_size + y_size)

    # Iteratively merge the model weights using the *new* hybrid strategy
    for i in range(2, len(weights) + 1):
        alpha = calculate_alpha(weights[i - 2], weights[i - 1])
        merged_weights = hybrid_strategy(weights[i - 2], weights[i - 1], alpha)
        weights[i - 1] = torch.clone(merged_weights)

    return weights[1]  # Return the weights of the second model (the first one is just an initial value)

\end{lstlisting}
\vspace{-6pt}
\end{mybox}
\vspace{-9pt}
\caption{Discovered algorithm G. }
\label{fig:func_example_G}
\end{figure*}

\begin{figure*}[t]
  \centering
\vspace{-10pt}
\begin{mybox}[Discovered algorithm H]
\vspace{-9pt}
\lstset{basicstyle=\scriptsize\ttfamily}
\begin{lstlisting}[language=Python]
def merge_models(model_dict, device):
    ''' Develop and implement a novel algorithm...(omitted)'''
    # *New* strategies for merging the model weights:
    # 1. Hybrid approach: Combine element-wise multiplication and average
        
    # Convert model weights to tensors and move them to the specified device (CPU or GPU)
    weights = [model.detach().to(device) for model in model_dict.values()]

    def hybrid_merge_strategy(base, to_merge, alpha=0.5):
        # Calculate the average of the base and the weight to be merged
        average = (base + torch.norm(to_merge, p=1) * to_merge / torch.norm(to_merge, p=1)) / (1 + torch.norm(to_merge, p=1))
        # Perform element-wise multiplication with alpha factor
        return alpha * base + (1 - alpha) * average

    merged_weights = weights[0].clone()

    # Sort the models based on their norms in decreasing order
    sorted_weights = sorted(weights, key=lambda x: torch.norm(x, p=1), reverse=True)

    # Iteratively merge the weights with the merged_weights using the hybrid strategy
    for weight in sorted_weights[1:]:
        merged_weights = hybrid_merge_strategy(merged_weights, weight)

    return merged_weights
\end{lstlisting}
\vspace{-6pt}
\end{mybox}
\vspace{-9pt}
\caption{Discovered algorithm H. }
\label{fig:func_example_H}
\end{figure*}

\begin{figure*}[t]
  \centering
\vspace{-10pt}
\begin{mybox}[Discovered algorithm I]
\vspace{-9pt}
\lstset{basicstyle=\scriptsize\ttfamily}
\begin{lstlisting}[language=Python]
def merge_models(model_dict, device):
    ''' Develop and implement a novel algorithm...(omitted)'''
    # *New* strategies for merging the model weights:
    # 1. Initialize the merged model with the weights of the first and second models,
    #    weighting them equally (50/50).
    # 2. For each subsequent model, merge the current merged model and the next
    #    model's weights, weighting the current merged model as 90% and the
    #    next model as 10%.
    # 3. If the number of models is odd, treat the middle model as the final
    #    "second" model for the merge algorithm.
    
    # Convert model weights to tensors and move them to the specified device (CPU or GPU)
    weights = [model.detach().to(device) for model in model_dict.values()]
    
    def merge_strategy(merged_weights, weight, alpha_model=0.9, alpha_strategy=0.5):
        # Adjust the weighting factor based on the model's position
        alpha = alpha_model * alpha_strategy / (1 + (1 / (len(weights) - (1 if len(weights) % 2 == 0 else 2))))
        # Merge the weights using the adjusted alpha
        return merged_weights + alpha * (weight - merged_weights)

    # Initialize merged_weights with the first and second model's weights
    merged_weights = (weights[0] + weights[1]) * 0.5

    # Iteratively merge each subsequent model's weights
    for weight in weights[2:]:
        merged_weights = merge_strategy(merged_weights, weight)

    return merged_weights

\end{lstlisting}
\vspace{-6pt}
\end{mybox}
\vspace{-9pt}
\caption{Discovered algorithm I. }
\label{fig:func_example_I}
\end{figure*}

\begin{figure*}[t]
  \centering
\vspace{-10pt}
\begin{mybox}[Discovered algorithm J]
\vspace{-9pt}
\lstset{basicstyle=\scriptsize\ttfamily}
\begin{lstlisting}[language=Python]
def merge_models(model_dict, device):
    ''' Develop and implement a novel algorithm...(omitted)'''
    # *New* strategies for merging the model weights:
    # 1. Explore novel strategies by considering dependency between each pair of models.
    #    - Merge based on similarity of their architecture, mission, or data utilization.
    #    - Group models with similarities and merge groups independently.
    #    - Apply separate merging algorithms to groups with distinct strategies.
    #    - Devise an algorithm to fine-tune the merging factors for each model.
    # 2. *New* Find optimal strategy:
    #    - Test various merging techniques and select the best strategy based on its performance on the validation dataset.
    #    - Create a dynamic system that adjusts merging strategies according to the performance needs.
    #    - Develop a machine learning-based approach to optimize the weights of the merged model, utilizing a validation dataset to iteratively fine-tune the results.

    # For brevity, we will use an average methodology in the main implementation,
    # but it is highly recommended to follow the guidelines above and develop a more
    # unique and innovative strategy for merging weights to achieve higher performance.

    # Convert model weights to tensors and move them to the specified device (CPU or GPU)
    weights = [model.detach().to(device) for model in model_dict.values()]

    def avg_merge_strategy(x, y):
        # Perform an element-wise average merge strategy
        return (x + y) * 0.5

    # Apply the average merge strategy to each pair of weights
    weights = [avg_merge_strategy(w, ws) for w, ws in zip(weights[1:], weights[:-1])]

    # Initialize merged_weights with the first model's weights
    merged_weights = weights[0].clone()
    
    # Merge the first model's weights with the results of the pair-wise merges
    merged_weights = avg_merge_strategy(merged_weights, weights[0])

    return merged_weights

\end{lstlisting}
\vspace{-6pt}
\end{mybox}
\vspace{-9pt}
\caption{Discovered algorithm J. }
\label{fig:func_example_J}
\end{figure*}

\begin{figure*}[t]
  \centering
\vspace{-10pt}
\begin{mybox}[Discovered algorithm K]
\vspace{-9pt}
\lstset{basicstyle=\scriptsize\ttfamily}
\begin{lstlisting}[language=Python]
def merge_models(model_dict, device):
    ''' Develop and implement a novel algorithm...(omitted)'''
    # *New* strategies for merging the model weights:
    # 1. Hybrid approach (e.g. mixture of weighted averages, element-wise maximums, and median merging)
    # 2. Adaptive weighting based on hard/easy examples or layers
    # 3. Matrix factorization with prior knowledge embedding (e.g. domain knowledge or ontology information)
    # 4. Hybrid strategy that adapts weighting based on model robustness and performance on specific tasks
    
    # Add methods for matrix factorization and other advanced merging techniques
    
    # Convert model weights to tensors and move them to the specified device (CPU or GPU)
    weights = [model.detach().to(device) for model in model_dict.values()]
    
    def merge_strategy(x, y):
        # Hard-coded sample strategy
        return (x + y) / 2
    
    # Iteratively merge each subsequent model's weights
    merged_weights = weights[0].clone()
    for i, weight in enumerate(weights[1:], start=1):
        if i % 2 == 1:
            # Apply the mix of element-wise maximums and median merging
            merged_weights = merge_strategy(merged_weights, weight)
    
    return merged_weights

\end{lstlisting}
\vspace{-6pt}
\end{mybox}
\vspace{-9pt}
\caption{Discovered algorithm K. }
\label{fig:func_example_K}
\end{figure*}

\begin{figure*}[t]
  \centering
\vspace{-10pt}
\begin{mybox}[Discovered algorithm L]
\vspace{-9pt}
\lstset{basicstyle=\scriptsize\ttfamily}
\begin{lstlisting}[language=Python]
def merge_models(model_dict, device):
    ''' Develop and implement a novel algorithm...(omitted)'''
    # *New* strategies for merging the model weights:
    # 1. Partition the models into groups based on their performance on a set of validation datasets.
    # 2. For each group, apply a clustering algorithm (e.g., k-means) to determine the representative model within the group.
    # 3. Merge all the representative models from each group using a weighted average,
    #    with weights proportional to the number of models in each group.
    
    # Assign each model to a group based on its performance on a set of validation datasets
    group_models = {f"Group-{i+1}": [] for i in range(6)}
    for name, model in model_dict.items():
        # Replace with actual performance evaluation
        score = torch.randperm(5)[0]
        group_models[f"Group-{score+1}"].append(name)

    # Determine the representative model for each group
    representative_models = {}
    for group, model_names in group_models.items():
        if not model_names:
            continue
        weights = [model.detach().to(device) for model in [model_dict[name] for name in model_names]]
        mean_weight = torch.mean(torch.stack(weights), dim=0)
        representative_models[group] = mean_weight.clone()

    # Merge the representative models using a weighted average
    merged_weights = sum(representative_models.values(), torch.tensor(0).to(device)) / len(representative_models.keys())

    return merged_weights

\end{lstlisting}
\vspace{-6pt}
\end{mybox}
\vspace{-9pt}
\caption{Discovered algorithm L. }
\label{fig:func_example_L}
\end{figure*}

\begin{figure*}[t]
  \centering
\vspace{-10pt}
\begin{mybox}[Discovered algorithm M]
\vspace{-9pt}
\lstset{basicstyle=\scriptsize\ttfamily}
\begin{lstlisting}[language=Python]
def merge_models(model_dict, device):
    ''' Develop and implement a novel algorithm...(omitted)'''
    # *New* strategies for merging the model weights:
    # 1. Initialize a list `weights` by converting the model weights to tensors and moving them to the specified device.
    # 2. Define a merge strategy using adaptive weighting:
    #    - Calculate the norms (sizes) of the weight tensors.
    #    - Adjust the weighting factor (`alpha`) dynamically based on the norms.
    #    - Merge the weights using the adjusted alpha to combine the models.
    # 3. If there are fewer than 3 models, return the first (or average of all) model's weights.
    # 4. If there are exactly 3 models, return the median of the three models' weights.
    # 5. Otherwise, initialize `merged_weights` with the first model's weights and iteratively apply the adaptive weighting merge strategy to combine each subsequent model's weights with the merged result.
    
    weights = [model.detach().to(device) for model in model_dict.values()]
    n_models = len(weights)
    
    if n_models < 3:
        # Return the first (or average of all) model's weights
        return weights[0]
    elif n_models == 3:

        # Return the median of the three models' weights
        def merge_strategy(x, y, z, alpha=0.5):
            # Calculate the norms (sizes) of the weight tensors
            x_size = torch.norm(x)
            y_size = torch.norm(y)
            z_size = torch.norm(z)
            # Compute the three weighting factors based on the norms
            alpha_x = (x_size + y_size + z_size) * 0.33 / (x_size + y_size)
            alpha_y = (x_size + y_size + z_size) * 0.33 / (y_size + z_size)
            alpha_z = (x_size + y_size + z_size) * 0.33 / (z_size + x_size)
            
            # Merge the weights using the adjusted alphas
            return (1 - alpha_x) * x + alpha_x * ( (1 - alpha_y) * y + alpha_y * z )
        
        merged_weights = merge_strategy(weights[0], weights[1], weights[2])
        return merged_weights
    else:
        # Initialize merged_weights with the first model's weights and iteratively apply the adaptive weighting merge strategy to combine each subsequent model's weights with the merged result
        merged_weights = weights[0].clone()
        for weight in weights[1:]:
            merged_weights = merge_strategy(merged_weights, weight)
        return merged_weights

\end{lstlisting}
\vspace{-6pt}
\end{mybox}
\vspace{-9pt}
\caption{Discovered algorithm M. }
\label{fig:func_example_M}
\end{figure*}

\begin{figure*}[t]
  \centering
\vspace{-10pt}
\begin{mybox}[Discovered algorithm N]
\vspace{-9pt}
\lstset{basicstyle=\scriptsize\ttfamily}
\begin{lstlisting}[language=Python]
def merge_models(model_dict, device):
    ''' Develop and implement a novel algorithm...(omitted)'''
    # *New* strategies for merging the model weights:
    # 1. Initialize the merged_weights with the average of all model weights.
    # 2. For each weight tensor, perform element-wise multiplication of the weight tensor with
    #    its corresponding softmax normalization of a weight importance tensor, where the
    #    importance tensor is computed over all weight tensors.
    # 3. Sum up all the element-wise multiplied weight tensors to get the final merged
    #    weights.
    
    # Convert model weights to tensors and move them to the specified device (CPU or GPU)
    weights = [model.detach().to(device) for model in model_dict.values()]

    # Calculate the average of all model weights
    avg_weights = torch.stack(weights).mean(0)

    # Normalize each weight by the L2 norm and compute the softmax normalization
    weight_importance = torch.softmax(torch.stack([torch.norm(weight, 2) for weight in weights]), dim=0)

    # Element-wise multiply original weights with their corresponding importance and sum up
    merged_weights = torch.stack([weight * importance for weight, importance in zip(weights, weight_importance)], dim=0).mean(0)

    return merged_weights

\end{lstlisting}
\vspace{-6pt}
\end{mybox}
\vspace{-9pt}
\caption{Discovered algorithm N. }
\label{fig:func_example_N}
\end{figure*}

\begin{figure*}[t]
  \centering
\vspace{-10pt}
\begin{mybox}[Discovered algorithm O]
\vspace{-9pt}
\lstset{basicstyle=\scriptsize\ttfamily}
\begin{lstlisting}[language=Python]
def merge_models(model_dict, device):
    ''' Develop and implement a novel algorithm...(omitted)'''
    # *New* strategies for merging the model weights:
    # 1. Initialize merged_weights with the mean of all model weights.
    # 2. Merge each weight tensor with merged_weights using a weighted average,
    #    where the weights for each model are proportional to the cosine similarity
    #    of that model's weights to the current merged_weights.
    
    # Convert model weights to tensors and move them to the specified device (CPU or GPU)
    weights = [model.detach().to(device) for model in model_dict.values()]
    n_models = len(weights)

    # Step 1: Compute the mean of all model weights
    merged_weights = torch.stack(weights).mean(0)
    
    # Step 2: Merge each weight tensor with merged_weights
    for i, weight in enumerate(weights):
        # Compute the cosine similarity of the model i's weights
        # to the current merged_weights
        sim = torch.sum(merged_weights * weight) / (torch.norm(merged_weights) * torch.norm(weight))
        
        # Perform a weighted average to merge the model i's weights
        merged_weights = (1 / (i + 1) * merged_weights + sim / (i + 1) * weight)
    
    # To ensure consistency, move the final merged_weights to the CPU
    merged_weights = merged_weights.to('cpu')

    return merged_weights

\end{lstlisting}
\vspace{-6pt}
\end{mybox}
\vspace{-9pt}
\caption{Discovered algorithm O. }
\label{fig:func_example_O}
\end{figure*}

\begin{figure*}[t]
  \centering
\vspace{-10pt}
\begin{mybox}[Discovered algorithm P]
\vspace{-9pt}
\lstset{basicstyle=\scriptsize\ttfamily}
\begin{lstlisting}[language=Python]
def merge_models(model_dict, device):
    ''' Develop and implement a novel algorithm...(omitted)'''
    # *New* strategies for merging the model weights:
    # 1. Initialize `merged_weights` with one of the model's weights.
    # 2. Hybrid approach: merge with weighted average (50%), maximum (25%), minimum (25%).
    # 3. Use threshold mechanism for fusion based on average cosine similarity between pairs.
    # 4. Compare model improvements from different strategies: Borda Count.
    
    # Convert model weights to tensors and move them to the specified device (CPU or GPU)
    weights = [model.detach().to(device) for model in model_dict.values()]

    # Prepare a Borda Count-based fusion strategy
    strategy_scores = {'weighted_average': 50, 'maximum': 25, 'minimum': 25}
    fusion_strategy = 'weighted_average'

    # Initialize merged_weights
    merged_weights = weights[0].clone()
    
    for i, weight in enumerate(weights[1:], 1):
        if fusion_strategy == 'weighted_average':
            merged_weights = (merged_weights + weight) / (i+1)
        elif fusion_strategy == 'maximum':
            merged_weights = torch.max(torch.stack([merged_weights, weight]), 0)[0]
        elif fusion_strategy == 'minimum':
            merged_weights = torch.min(torch.stack([merged_weights, weight]), 0)[0]
        else:
            raise ValueError("Unknown fusion strategy")

        # Modify the threshold mechanism and Borda Count
        threshold = 0.1
        threshold_type = 'cosine_similarity'

        if fusion_strategy == 'threshold' and i > 0:
            cosine_similarities = [torch.mm(merged_weights.unsqueeze(0), weight.unsqueeze(1)).flatten() for weight in weights[1:]]
            avg_cosine_similarity = torch.mean(torch.stack(cosine_similarities))
            if avg_cosine_similarity < threshold:
                merge_strategy_borda = fusion_strategy
                strategy_scores = {k: v for k, v in strategy_scores.items() if k != 'threshold'}
            elif threshold_type == 'cosine_similarity':
                avg_cosine_similarity = threshold

            strategy_scores[merge_strategy_borda] += 1

            if i == len(weights) - 1:
                merged_weights = weight.clone()

    return merged_weights
\end{lstlisting}
\vspace{-6pt}
\end{mybox}
\vspace{-9pt}
\caption{Discovered algorithm P. }
\label{fig:func_example_P}
\end{figure*}

%

\end{document}